\documentclass[acmtog,timestamp]{acmart}
\pdfoutput=1
\acmSubmissionID{288}

\usepackage{amsmath,amsthm}
\usepackage{amssymb}
\usepackage{graphicx}
\usepackage{textcomp}
\usepackage{wrapfig}
\usepackage{subfig}
\usepackage{color}
\usepackage{xspace}
\usepackage{overpic}
\usepackage{subfig}
\usepackage{enumitem}
\usepackage{booktabs}
\usepackage{algorithm2e}
\usepackage{hyperref}
\usepackage{comment}
\usepackage{multirow}

\definecolor{blue}{rgb}{0,0,1}
\definecolor{red}{rgb}{1,0,0}
\definecolor{green}{rgb}{0,.5,0}
\definecolor{orange}{rgb}{0.75, 0.4, 0}
\definecolor{teal}{rgb}{0.0, 0.4, 0.4}
\definecolor{purple}{rgb}{0.65,0,0.65}

\newcommand{\rz}[1]{{\color{black}\textbf{}#1}\normalfont}
\newcommand{\rh}[1]{{\color{black}\textbf{}#1}\normalfont}
\newcommand{\lm}[1]{{\color{black}\textbf{}#1}\normalfont}

\renewcommand{\paragraph}[1]{\textbf{#1}}

\begin{document}

\title{Learning High-DOF Reaching-and-Grasping via Dynamic Representation of Gripper-Object Interaction}

\author{Qijin She}
\authornote{Joint first authors.}
\email{qijinshe@outlook.com}
\affiliation{
 \institution{National University of Defense Technology}
 \country{China}
}
\author{Ruizhen Hu}
\authornotemark[1]
\email{ruizhen.hu@gmail.com}
\affiliation{
 \institution{Shenzhen University}
 \country{China}
}

\author{Juzhan Xu}
\email{juzhan.xu@gmail.com}
\affiliation{
 \institution{Shenzhen University}
 \country{China}
}

\author{Min Liu}
\email{gfsliumin@gmail.com}
\affiliation{%
 \institution{Chinese Academy of Military Science}
 \country{China}
}

\author{Kai Xu}
\authornote{Corresponding authors.}
\email{kevin.kai.xu@gmail.com}
\affiliation{%
 \institution{National University of Defense Technology}
 \country{China}
}

\author{Hui Huang}
\authornotemark[2]
\email{hhzhiyan@gmail.com}
\affiliation{%
 \institution{Shenzhen University}
 \country{China}
}

\begin{teaserfigure}
	\centering
	\includegraphics[width= \linewidth]{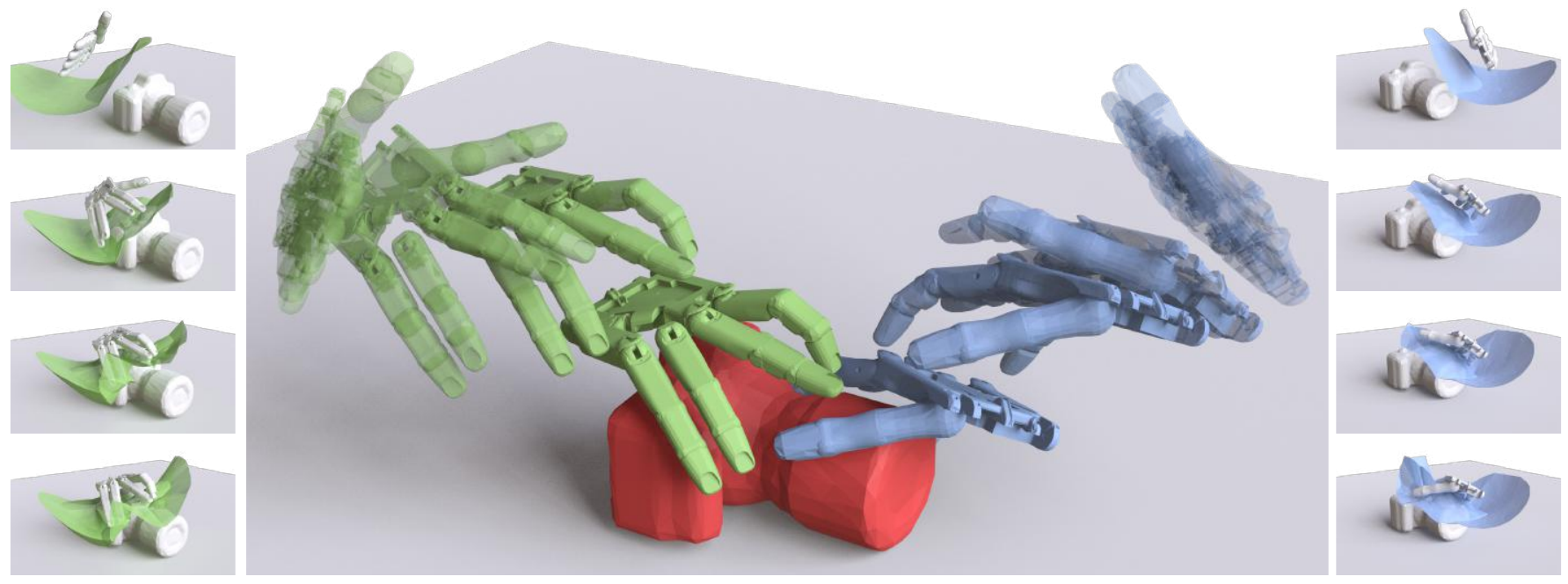}
	\caption{Results of our reaching-and-grasping motion planning method for the red camera put on the table given two different initial configurations of the gripper shown in green and blue colors, respectively. For each result, we show four key frames of the reaching process and final grasping pose in the middle and show the corresponding Interaction Bisector Surfaces (IBSs) used to encode the interactions on the side. Note that we enlarge the distance between the gripper and the camera for the frames shown in the middle to show the gripper configurations more clearly. }
		\label{fig:teaser}
\end{teaserfigure}

\begin{abstract}
\rz{We approach the problem of high-DOF reaching-and-grasping via learning joint planning of grasp and motion with deep reinforcement learning. 
To resolve the sample efficiency issue in learning the high-dimensional and complex control of dexterous grasping, 
we propose an effective representation of grasping state characterizing the spatial interaction between the gripper and the target object. 
To represent gripper-object interaction, we adopt Interaction Bisector Surface (IBS) which is the Voronoi diagram between two close by 3D geometric objects 
and has been successfully applied in characterizing spatial relations between 3D objects. We found that IBS is surprisingly effective as a state representation 
since it well informs the fine-grained control of each finger with spatial relation against the target object. 
This novel grasp representation, together with several technical contributions including a fast IBS approximation, 
a novel vector-based reward and an effective training strategy, facilitate learning a strong control model of high-DOF grasping with good sample efficiency, 
dynamic adaptability, and cross-category generality. Experiments show that it generates high-quality dexterous grasp for complex shapes with smooth grasping motions.}
\end{abstract}


%

\maketitle

\section{Introduction}

Robotic grasping is an important and long-standing problem in robotics. It has been drawing increasingly broader attention from the fields of computer vision~\cite{saxena2010vision}, machine learning~\cite{kleeberger2020survey} and computer graphics~\cite{liu2009dextrous,pollard2005physically}, due to the interdisciplinary nature of its core techniques.
Traditional methods typically approach grasping by breaking the task into two stages: static grasp synthesis followed by motion planning of the gripper. In the first stage, a few candidate grasp poses are generated. The second stage then plans a collision-avoiding trajectory for the robotic arm and gripper to select a feasible grasp and execute it. The main limitation of such a decoupled approach is that the two stages are not jointly optimized which could lead to suboptimal solutions.

An integrated solution to grasp planning and motion planning, \rz{which is often referred to as the \emph{reach-and-grasp problem}}, remains a challenge~\cite{wang2019manipulation}.
A few works have attempted integrated grasp and motion planning by formulating a joint optimization problem. The advantage of an integrated planner is that motion planning and grasp planning impose constraints on each other. However, these existing methods still rely on pre-sampled grasp candidates over which a probabilistic distribution for selection is computed, making it highly reliant on the quality of the candidates. Wang et al.~\shortcite{wang2019manipulation} introduce online grasp synthesis to eliminate the need for a perfect grasp set
and grasp selection heuristics. Nevertheless, such an approach optimizes over a discrete set of grasp candidates, which limits the grasping space explored.

Reinforcement learning (RL)~\cite{sutton2018reinforcement} models offer a counterpoint to the planning paradigm. Rather than optimizing for grasp selection and motion planning, the idea is to use closed-loop feedback control based on sensory observations so that the agent can dynamically update its strategy while accumulating new observations. More recent advances of RL allow for continuous, high-dimensional actions which are especially suitable for continuous exploration of \rz{reach-and-grasp} planning. Albeit offering promising solutions, the sample efficiency issue of RL hinders its application in highly complex control scenarios, such as dexterous grasping of a high-DOF robotic hand (e.g., a 24-DOF five-fingered gripper).

We argue that the main cause of the limitation above is the lack of an \rz{effective} representation of observations. Indeed, even with deep neural networks as powerful function approximators, it is still too difficult to fit a function mapping raw
sensory observations (camera images) to low-level robot actions (e.g., motor torques, velocities, or Cartesian motions).
Therefore, learning complicated control of high-DOF grippers calls for an informative representation of \rz{the intermediate states during the reaching-and-grasping process.} 
Such representation should well inform the RL model about \rz{the \emph{dynamic interaction} between the gripper and the target object.}

In this work, we advocate the use of \emph{Interaction Bisector Surface (IBS)} for representing gripper-object interaction in RL-based \rz{reach-and-}grasp learning. IBS, computed as the Voronoi diagram between two geometric objects, was originally proposed for indexing and recognizing inter-protein relation in the area of biology~\cite{kim2006interaction}. In computer graphics, it has been successfully applied in characterizing fine-grained spatial relations between 3D objects~\cite{zhao2014IBS,hu2015interaction}.
We found that IBS is surprisingly effective as a state/observation representation in learning high-DOF \rz{reach-and-}grasp planning.
Gripper-object IBS well informs \rz{the global pose of the gripper and }the fine-grained \rz{local} control of each finger with spatial relation against the target object, making the map from observation to action easier to model.
\rz{For different initial configuration of the gripper, i.e., different relative pose to the object, our method is able to provide different motion sequences and form different final graspings as shown in Figure~\ref{fig:teaser}.
Moreover, during the reaching-and-grasping process, the dynamic change of  relationship between the gripper and object can be well-reflected by the corresponding IBS, which enables our method to deal with moving objects with dynamic pose change, going beyond static object grasping of most previous works.
In addition, as IBS is defined purely based on the geometry of the gripper and the object without any assumption of the object semantics or affordance, our method can generalize well to objects of unseen categories.

To speed up the computation of IBS for efficient training and testing, we propose a grid-based approximation of IBS as well as post-refinement mechanism to improve accuracy. Empirical studies show that the approximation is accurate to capture the important interaction information and fast enough to enable online computation in an interactive frame rate.
To capture richer information of interaction, we propose a combination of local and global encoders for multi-level feature extraction of IBS, based on its segmentation corresponding to the different components of the gripper.
}

We adopt Soft Actor-Critic (SAC)~\cite{haarnoja2018soft}, an off-policy model, as our RL framework.
Aside from the core design of state representation, we introduce two other critical designs to make our model more sample efficient and easier to train. To learn collision-avoiding finger motions, we impose finger-object contact information as a constraint of RL. A straightforward option is to design the reward to punish finger-object penetration. This can greatly complicate model training. Thus, our \emph{first key design} is to enhance the standard scalar Q value into a vector storing finger-wise contact information. Such disentangled Q representation provides a more informative dictation on learning contact-free motions.

To deal with the continuous action space in grasp planning, 
\rh{we opt to learn from imperfect demonstrations synthesized offline with heuristic planning from different initial configurations to the final grasp poses generated GraspIt!~\cite{miller2004graspit}. }
The demonstration grasping trajectories, possibly containing gripper-object penetration, 
are stored in the experience replay buffer~\cite{mnih2015human} of SAC. 
Our \emph{second key design} is to bootstrap the learning with a bootstrapping replay buffer containing imperfect demonstrations possibly with collision. 
We then inject the replay buffer with an increasing amount of experiences sampled from the currently learned policy and rectified to be collision-free. 
This forms an enhanced replay buffer based on which a gradually refined policy can be learned. 
Such a double replay buffer scheme helps learn a strong control planning model fairly efficiently.

\rz{Our method generates high-quality dexterous grasps for unseen complex shapes with smooth grasping motions.
	 Furthermore, our method can dynamically adjust and adapt to object movement during grasping, thus allowing to grasp moving objects.
	When compared to other baseline methods, our method consistently achieves a higher success rate on different datasets.
Our method is also robust to partial observations of target objects.  }
Our contributions include:
\begin{itemize}
  \item A novel state representation for learning \rz{reach-and-grasp} planning based on gripper-object interaction bisector surface, along with an accurate approximation for fast training.
  \item A combination of local and global encoders for multi-level feature extraction of IBS to capture richer information of gripper-object interaction.
  \item A new vector-based representation of Q value encoding not only regular rewards but also finger-wise contact information for efficient learning of contact-free grasping motion.
  \item A double replay buffer mechanism for learning collision-avoiding grasping policy from imperfect demonstrations.
\end{itemize}

\section{Related Work}
Robotic grasping has a large body of literature. Existing approaches can generally be classified into analytical and data-driven methods.
Analytical (or geometric) approaches analyze the shape of the target object to synthesize a suitable grasp~\cite{sahbani2012overview}. Data-driven (or empirical) approaches based on machine learning are gaining increasing attention in recent years~\cite{bohg2013data,kleeberger2020survey} .


\begin{figure*}[!t]
    \centering
    \includegraphics[width=0.96\textwidth]{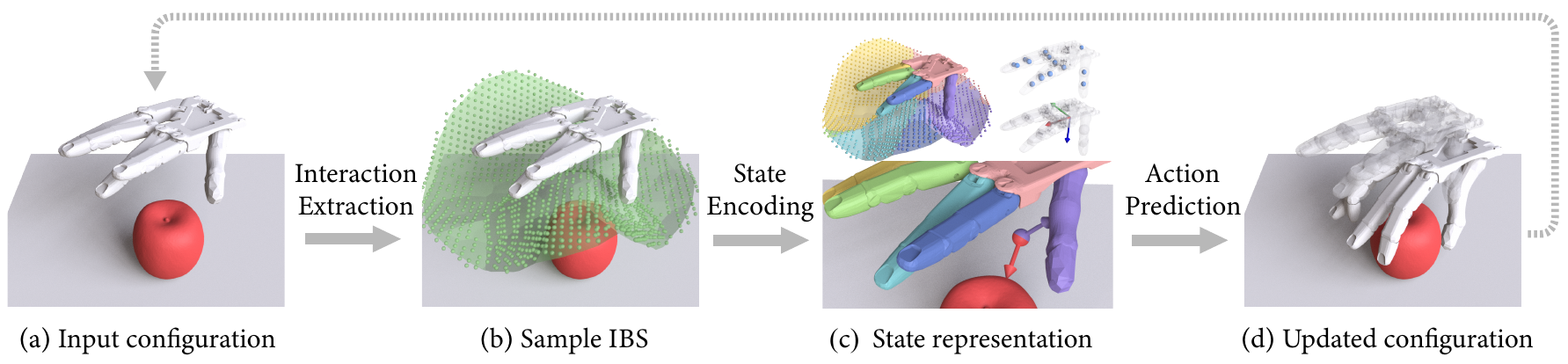}
\caption{
Overview of one iteration of our grasping motion planning method. Given an object in a scene context with the current gripper configuration, our method first generates the sampled IBS to represent the interaction between the scene and gripper, then a set of local and global features are extracted from the given state to predict the action that changes the configuration of the gripper so that it moves closer to the given object and forms a better grasping. The updated configuration after applying the predicted action is then passed through the same pipeline to predict the subsequent action.}
\label{fig:overview}
\end{figure*}

\rz{\paragraph{Analytical robotic grasping.}
With known object shapes, analytical methods \lm{search for grasp poses that maximize a certain grasp quality metric, and these methods can mainly be classified into discrete sampling-based techniques or continuous optimization techniques. Some sampling-based techniques search in the space of grasp poses~\cite{miller2000graspit,miller2004graspit}, while others search contact points or contact areas on surfaces of objects and then searches for collision-free grasp poses that realize the given set of contact points or areas~\cite{chen1993finding,7812687,9120282,9626457}. Compared to sampling-based techniques, continuous optimization techniques plan grasp poses by optimizing differentiable losses~\cite{kiatos2019grasping,maldonado2010robotic}, which are more efficient. Recent work~\cite{liu2020deep} proposes a differentiable grasp quality, which can be used for continuous optimization and deep learning methods.
For high-DOF grippers, however, sampling-based techniques are computationally costly due to the large search space. Continuous optimization techniques for high-DOF grasp planning apt to find local optimal grasp poses.} 
  Moreover, analytical methods are difficult to generalize to incomplete or unknown objects.
}

\paragraph{Learning-based robotic grasping.}
Learning-based methods are typically split into supervised learning and reinforcement learning.
For supervised learning, grasp annotations can be collected either by humans~\cite{depierre2018jacquard}, with simulation~\cite{mahler2016dex}, or through real robot tests~\cite{levine2018learning}. Supervised grasp learning can be categorized as discriminative or generative depending on whether the grasp configuration is the input or output of the learned model.
While discriminative approaches sample grasp candidates and rank them using a neural network~\cite{mahler2018dex},
generative approaches directly generate suitable grasp poses~\cite{morrison2018closing}.

Early learning-based works mainly focus on generating grasps for target objects with low-DOF grippers (such as parallel-jaw grippers)~\cite{saxena2007robotic,gualtieri2016high}. Such work learns to regress grasp quality or to predict grasp success ~\cite{mahler2016dex,inproceedingsDexNetTwo,fang2018mtda,lu2020multifingered,lu2020planning,van2019learning}, but still needs sampling-based techniques to search for better grasps. Liu et al.~\shortcite{liu2019grasp,liu2020deep} use deep neural networks to directly regress high-DOF grasps based on the input of voxels or depth images.

\paragraph{Reinforcement learning for robotic grasping.}
Deep reinforcement learning (RL) has been shown as a promising and powerful technique to automatically learn control policies by trial and error. Based on raw sensory inputs, dexterous grasping behaviors can be performed. A comparative study of RL-based grasping methods is given in~\cite{quillen2018deep}.
QT-Opt~\cite{kalashnikov2018qt} learns various manipulation strategies with dynamic responses to disturbances.
Song et al.~\shortcite{song2020grasping} present an RL-based closed-loop 6D grasping of novel objects with the help of human demonstrations. The learned policy can operate in dynamic scenes with moving objects.
Rajeswaran et al.~\shortcite{rajeswaran2017learning} show that model-free RL can effectively scale up to complex
manipulation tasks with a high-DOF hand.
Mandikal and Grauman~\shortcite{mandikal2020dexterous} introduce an approach for learning dexterous grasps, and the key idea is to embed an object-centric visual affordance model within a deep reinforcement learning
loop to learn grasping policies that favor the same object regions favored by people. \rz{Andrychowicz et al.~\shortcite{andrychowicz2020learning} explore RL for dexterous in-hand manipulation by reorientating a block. Starke et al.~\shortcite{starke2019neural} present a good example of conditioning motion behavior on the geometry of the surrounding 3D environment. Ficuciello et al. propose methods that use RL to search good grasp poses \shortcite{ficuciello2016synergy} as well as good grasp motion trajectories \shortcite{monforte2019multifunctional} in a synergies subspace, and these methods need good initial parameters to reduce the searching space while finding initial parameters need additional imitation learning or human efforts.
To overcome this limitation, Ficuciello et al.~\shortcite{ficuciello2019vision} introduce a visual module to predict initial parameters given visual information of objects. The main weakness of their methods is that the learning process should be done for every object respectively to achieve a good grasp.
}

Training RL models in the real environment is usually prohibitive since it requires a large number of trials and errors.
A straightforward approach is to train them in a simulation environment and then transfer the learned policy to the real world with sim-to-real techniques~\cite{peng2018sim,james2019sim}.
Our work trains an RL model for dexterous high-DOF grasping.
We focus on how to train such complicated 
planning policies with the help of an effective representation of gripper-object interaction
and leave the issue of sim-to-real transfer for future work.


\paragraph{Grasp representation.}
Existing grasping models have adopted various representations to describe the shape of the target object to be grasped, such as voxels~\cite{varley2017shape}, depth images~\cite{viereck2017learning}, multi-view images~\cite{collet2010efficient}, or geometric primitives~\cite{aleotti20123d}.
Some works represent a grasp with Independent Contact Regions (ICRs)~\cite{roa2009computation,fontanals2014integrated}.
These regions are defined such that if each finger is positioned on its corresponding contact region, a force-closure
grasp~\cite{nguyen1988constructing} is always obtained, independently of the exact location of each finger.

Our work opts to characterize the interaction between the gripper and the object using Interaction Bisector Surface (IBS)~\cite{zhao2014IBS}. Interaction Bisector Surface (IBS) captures the spatial boundary between two objects. It can provide a more detailed and informative interaction representation with both geometric and topological features extracted on the IBS. Hu et al.~\shortcite{hu2015interaction} further combined IBS with Interaction Region (IR), which
is used to describe the geometry of the surface region on the object corresponding to the interaction, to encode more
geometric features on the objects.
Pirk et al.~\shortcite{pirk2017understanding} build a spatial and temporal representation of
interactions named interaction landscapes.

\rh{Karunratanakul et al. ~\shortcite{karunratanakul2020grasping} proposes an implicit representation to encode hand-object grasps by defining the distance to human hand and objects for points in space, and then use it to generate the grasping hand pose for a given static shape.
In contrast, our method utilizes constantly updated IBS as state representation to plan the dynamic motion of the gripper for the object reach-and-grasp task.
}

\section{Overview}

Given the input scene segmented into foreground object and background context and the initial configuration of the gripper, our goal is to output a sequence of collision-free actions which move the gripper towards the object to perform a successful grasping. We train a deep reinforcement learning model in which the state is given by the gripper-object interaction and the actions are designed as the gripper configuration changes. Figure~\ref{fig:overview} gives an overview of one step of the planned reaching-and-grasping motion. 

Our method starts by extracting an informative representation of the interaction between the given object and the current gripper configuration. A set of local and global features are extracted from the given state to predict the action that changes the configuration of the gripper so that it moves closer to the given object and forms a better grasping. The updated configuration after applying the predicted action is then passed through the same pipeline to predict the subsequent action.

\paragraph{Interaction extraction.}
We use IBS~\cite{zhao2014IBS} to represent the interaction.
Since the computation of accurate IBS is time-consuming, we design an IBS sampler to obtain a discretized and simplified version of IBS in a much more efficient manner. The set of sampled IBS points is then moved to be closer to the exact IBS.

\paragraph{State encoding.}
\rz{The gripper configuration is encoded by its global position and orientation as well as local joint angels with (6+18)-DOF.
Based on the part components of the gripper model, i.e., five fingers and one palm, we adopt a multi-level representation of IBS inspired by the work of Zhao et al.~\shortcite{zhao2017character}.
More specifically, for each sampled IBS point, we first find its nearest points on
the scene and the gripper, and then encode it with a set of information, consisting of its own coordinate, the component labels of those two nearest points as well as the spatial relationship relative to each nearest point.
Therefore, the components of the gripper model, including five fingers and one palm, naturally induce a segmentation of the IBS based on the association between the gripper and the IBS. }
We then combine local features extracted separately for each IBS segment and global feature for the entire IBS to form a multi-level description of gripper-object interaction.


\paragraph{Action prediction.}
To predict the action given the current state, we design 
a policy network that takes both local and global features from the current configuration \rz{and outputs the configuration change of the gripper and terminate value.
The policy network is trained via reinforcement learning \rz{using the Soft Actor-Critic method~\cite{haarnoja2018soft}}.}
To avoid gripper-scene collision and self-collision of the gripper during the reaching-and-grasping process, 
we design \rz{a new vector-based representation of Q value encoding not only regular rewards but also finger-wise contact information for efficient learning of contact-free grasping motion.
To further accelerate the training, we generate a set of imperfect demonstration data to bootstrap the learning and to converge to the final policy in a more efficient and effective manner.}

\section{Method}

\begin{figure}[!t]
	\centering
	\includegraphics[width=\linewidth]{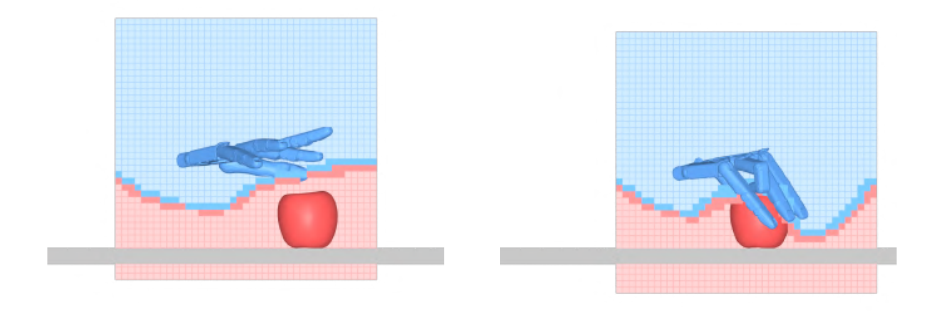}
	\caption{
		Examples of IBS sampled in two intermediate states when the gripper moves towards the object. For each state, we only show one vertical slice of the grids around the gripper to view the change of the IBS more clearly. Grids that are closer to the object are colored in red, and others are colored in blue. The IBS by definition is located on the boundary of those two colored regions, and we take the grid points in blue cells on the boundary to be the set of sampled IBS points.}
	\label{fig:ibs_sampler}
\end{figure}

\subsection{IBS sampler}
The IBS is essentially the set of points equidistant from two sets of points sampled on the scene and the gripper, respectively.
The computation of exact IBS requires the extraction of the Voronoi diagram, which is time-consuming. To trade-off between efficiency and accuracy, we compute IBS only within a given range and discretize the space to obtain an approximation of IBS, as shown in Figure~\ref{fig:ibs_sampler}.

More specifically, IBS is only computed within the sphere which is located at the \rz{centroid} of the palm with a radius of $r$. The bounding box of the sphere is discretized into a $k^3$ grid for IBS point sampling. For each of the grid points, i.e., the center of each grid cell, we compute its distances to both the scene $d_s$ and the gripper $d_g$, and the points having $\delta = d_g-d_s = 0$ are IBS points.
To find such set of points, we store the $\delta$ values on \rz{ grid cells} and extract the zero-crossing points on the fly.
\rz{Note that to accelerate the whole process, the $\delta$ values on only a partial set of cells around the exact IBS are computed in a region growing manner.
More specifically, we first find the cell  that is closest to the middle position of the line connecting the centroid of the foreground object in the scene and the palm, and compute its $\delta$ value.  Then the computation grows outwards with the neighboring cells until sufficient  grid cells that share the same sign have been found.}

\rz{Figure~\ref{fig:ibs_sampler} shows the sampling process of IBS in two different states. We can see that the grid is divided by IBS into two parts, i.e., one with points closer to the scene (in red color) and the other with points closer to the gripper (in blue color). Only the $\delta$ values on the grid cells close to the exact IBS are computed to get the initial set of sampled IBS points (highlighted with solid colors).
Note that such set of sampled IBS contains two layers of grid points which has redundant information, thus we only keep the layer corresponding to the gripper as our IBS point set (the solid layer in blue).}


\rz{Since the initial set of sampled IBS points are grid centers and the approximation accuracy is highly affected by the grid resolution, we further refine the locations of these points to make them approach the exact IBS.} 
For each point $p$, we first locate its nearest points on scene $p_s$ and gripper $p_g$. Without loss of generality, let us assume that the distance $d_g$ between the point to the gripper is larger than the distance $d_s$ between the point and the scene. We then need to move $p$ towards $p_g$ to make it closer to the exact IBS: $p' = p + \Delta_p \times \overrightarrow{pp_g}$. The detail of how to derive an appropriate $\Delta_p$ can be found in the supplementary material. Once the location of point $p$ is updated, we update its nearest points on the scene and on the gripper.

\begin{figure}[!t]
	\centering
	\includegraphics[width=0.94\linewidth]{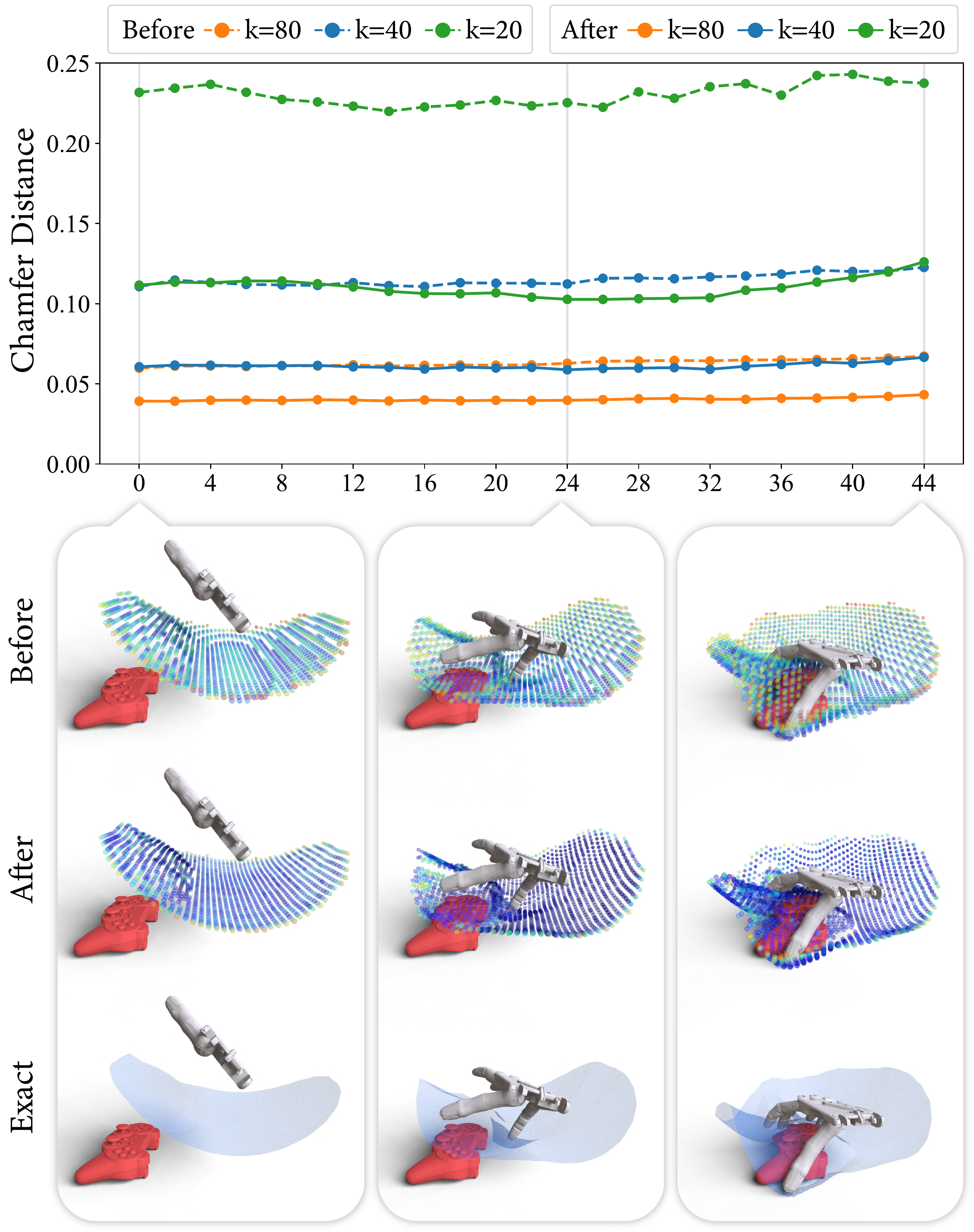}
	\caption{
		\rz{IBS approximation error during the reaching-and-grasping process before and after the refinement under different grid resolutions $k = \{20, 40, 80\}$.  The line chart shows how the approximation error changes during the process under different settings, and below we show the corresponding IBS samples before and after refinement compared to the exact IBS surface for three representative frames under $k=40$. We use the jet colormap on the  sampled points to indicate the approximation error, which shows that the approximation accuracy is highly improved after refinement.}
	}
	\label{fig:ibs_resolution}
\end{figure}

\rz{To gauge how close the sampled IBS is to the exact IBS as well as the effectiveness of our refinement method, we measure the IBS approximation error using the Chamfer distance between the points sampled from the grids and those sampled on the exact IBS surface.
Figure~\ref{fig:ibs_resolution} shows how the Chamfer distance changes during the reach-and-grasp process before and after the refinement under different grid resolutions $k = 20$, $40$, and $80$.
We can find several interesting properties.
Firstly,  approximation error is relatively stable during the reach-and-grasp process for each setting and gets slightly higher towards the end as the IBS surface becomes more complex.
Secondly, approximation error drops significantly after the refinement 
for different grid resolutions, which shows the effectiveness of the refinement process.
Thirdly, approximation error generally increases with the decrease of grid resolution, especially when no refinement is involved.
However, the computation time increases with the grid resolution. To obtain a good balance, we set grid resolution $k=40$ and sample $n=4096$ points for IBS approximation.}

\subsection{State and action representation}
\label{sec:state}
\paragraph{State representation.}
Finding an informative representation for a given state is the key to guiding the movement of the gripper towards a successful grasping of the object. Our key observation is that the IBS between the scene and gripper together with the gripper configuration provide rich information about the intermediate state.

For gripper configuration, we use a (6+18)-DOF Shadow Hand in all our experiments, where the first 6-DOF encodes the global orientation and position of the gripper and the remaining 18-DOF encodes the joint angles.
\rz{To better describe the local context around the gripper to guide its configuration change, we set the origin of the world coordinate to the centroid of the palm, as the setting in the IBS sampling process, to encode the spatial interaction features between the gripper and the scene.
We found that our method gets similar performance when using the centroid of the object as the origin of the world coordinate.
}

\begin{figure}[!t]
	\centering
	\includegraphics[width=0.9\linewidth]{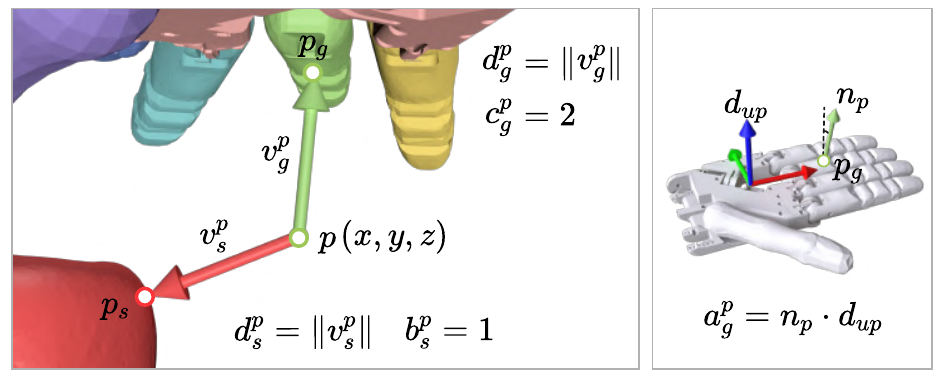}
	\caption{
	Informative representation of each sampled IBS point $p$, which includes a set of relative spatial information to either the scene or the gripper. A more detailed explanation of the notations can be found in Section~\ref{sec:state}. }
	\label{fig:ibs_encode}
\end{figure}

For each point $p$ on the sampled IBS, we store the following information as illustrated in  Figure~\ref{fig:ibs_encode}:
\begin{itemize}
	\item coordinate $ c = (x,y,z) \in  R^3$ 
	\item distance to the scene  $d_s^p \in  R$
	\item unit vector pointing to the nearest point $p_s$ on the scene $v_s ^p\in  R^3$
	\item indicator of whether $p_s$ is located on the foreground object $b_s^p \in \{0,1\}$
	\item distance to the gripper $d_g^p \in R$
	\item unit vector pointing to the nearest point $p_h$ on the gripper $v_g^p \in  R^3$
	\item\rz{one-hot indicator of the gripper component that $p_g$  belongs to $c_g^p \in  \{0, 1\}^6$ } 
	\item value defined on $p_g$ indicating which side of the gripper it located on $a_g^p \in [-1, 1]$
\end{itemize}
Here, $a_g^p = n_p \cdot d_{up}$ is the dot product of the normal direction $n_p$ of point $p_g$ on gripper in rest pose and the upright direction $d_{up}$ perpendicular to the palm and pointing outwards.

\paragraph{Action representation.}
\rz{The action consists of two parts. The first part} is defined as the gripper configuration change, which is also with (6+18)-DOF. For each single action, we restrain the change of each parameter within $[-0.25cm, 0.25cm]$ for global translation \rz{of the entire gripper}, and $[-0.025, 0.025]$ in radian for \rz{both global rotation angle of the entire gripper or local rotation angle of each joint.} 
\rz{The other part is the stop action, defined as a termination value to indicate the possibility of terminating the planning process.
The exact termination mechanism is provided in Section~\ref{sec:reward}.
.}

\subsection{Network and reward design}
\label{sec:reward}

\begin{figure}[!t]
	\centering
	\includegraphics[width=0.96\linewidth]{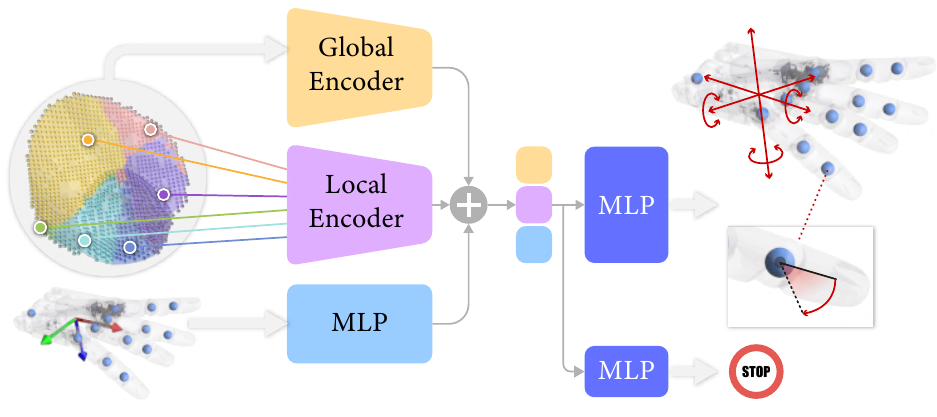}
	\caption{
		Grasp planner network used in our work. Given the rich information stored in the sampled IBS and current gripper configuration, we use both global and local encoders to extract multi-level features from the sample IBS and then concatenate them with the feature extracted from the gripper via MLP. The concatenated feature is passed to \rz{two other MLPs to get the final predicted action, one of which is defined as the global  translation and rotation of the gripper as well as the local joint rotation, and the other is defined as the termination probability.}
		}
	\label{fig:actor}
\end{figure}

\paragraph{Network architecture.}
Figure~\ref{fig:actor} shows the network architecture of our grasp planner (actor network).
For the sampled IBS, both local and global features are extracted to concatenate with the feature extracted from the gripper configuration, and then the concatenated feature is passed to \rz{ two other MLPs } to get the predicted action, \rz{one for the gripper configuration change and the other for the terminal value.}
We use PointNet~\cite{qi2017pointnet} for both local and global IBS encoder, and the global encoder takes the whole IBS as input while the local encoder takes each IBS component corresponding to different gripper components as input.

\paragraph{Reward function.}
\rz{
The reward function needs to reflect the quality of an executed action.
As our final goal is to perform a successful grasping of the given object in the scene at the end of the planning,
we first define a \emph{grasp reward} function $R_g$ to measure  the grasp quality when the whole process is terminated.
Moreover, to further encourage a more natural grasp pose with more gripper components taking part in the grasping and avoiding collision with the scene during the whole process, we define another \emph{reaching reward} function $R_c^i$ per gripper component $g_i$ to provide guidance for each intermediate step.

We measure the grasp quality from two aspects. First is the commonly used execution success, which we use the success signal $S$ obtained from the Pybullet simulator when performing the final grasp. 
\rh{We consider the grasp to be successful if and only if the object can be lifted up by more than 0.2m as in the previous work ~\cite{xu2021adagrasp}.}
More details about the simulator setup and how to perform the grasp can be found in the supplementary material. 
However, this sparse boolean value cannot provide enough guidance for high-DOF grasping, thus we complement it with another well-known geometric measure $Q_1$~\cite{ferrari1992planning}.
But as the traditional $Q_1$  measure can only be computed  when the gripper touches the object without any collision and the computation becomes unstable during the training, we adopt the generalized $Q_1$ measure proposed in~\cite{liu2020deep} instead. 


As those two grasp quality metrics are only computed when the reach-and-grasp task is completed, to encourage a quick convergence, we set a negative reward $r_f$ for each intermediate step.
So the grasp reward function is defined as:
	\[ R_g=
\begin{cases}
	\omega_s S + \omega_q Q_1, & \text{if the task is completed};\\
	r_f, & \text{otherwise}.
\end{cases} \]
where we set $\omega_s  = 150$, $\omega_q = 1000$, and $r_f = -3$  in all our experiments.
To encourage the contact between the gripper and the object, our planning process terminates requires that not only if a randomly sampled value is smaller than the terminal value but also at least two gripper components contact the object in the current step.

To determine whether a gripper component $g_i$ is contacting with the object, we would like to ensure that there are enough points on the gripper component that are close enough to the object while not colliding with the scene.
So we first count the number $m_i$ of IBS points determined together by the inner side of the gripper component $g_i$ ($c_n^p=i, a_g^p \geq 0$) and the object ($b_s^p = 1$) with distance to the object smaller than a given threshold $\delta_d = 0.5cm$ ($d_s^p < \delta_d $), i.e., IBS points with $b_s^p = 1, c_n^p=i, a_h^p \geq 0, d_s^p < \delta_d $.
Then we further check if any of the IBS points determined by the gripper component $g_i$  (i.e., IBS points with $c_n^p=i$) is on the inner side of the scene by computing the angle between the corresponding vector  $v_s^p$ pointing from the IBS point to the nearest point on the scene and the normal $n_s^p$ of the nearest point $p_s$. If the angle is smaller than $90$ degrees, we consider the IBS point is on the inner side of the scene, and count the number $n_i$ of such IBS points. If  $  n_i \geq \delta_n$, we consider the gripper component is colliding with the scene. 
Therefore, if  the gripper component $g_i$  is not colliding with the scene ($n_i <\delta_n $) and there are enough contacting points ($m_i \geq \delta_m $), we consider the gripper component is contacting the object.

Thus we define the reaching reward function $R_c^i$ per gripper component $g_i$ as follows to encourage a more effective reaching-and-grasping process with more contacting but not colliding points:}
\[ R_c^i=
\begin{cases}
	-100,  &  n_i \geq \delta_n (\text{collide  with the scene} );\\
	R_{\text{contact}}^i, & \text{otherwise}.
\end{cases} \]
\[ R_{\text{contact}}^i =
\begin{cases}
	\min\{\eta, m_i\},  & m_i \geq \delta_m (\text{have enough contact} );\\
	0, & \text{otherwise}.
\end{cases} \]
In all our experiments, we set $\delta_n = \delta_m = 3$ and $\eta = 40$.

\subsection{Network training}
\label{sec:training}
To train the network, we adopt the well-known off-policy method Soft Actor-Critic~\cite{haarnoja2018soft} and make some small modifications to the Q-network to make the training more effective.

SAC has two networks, i.e., the policy network (also known as the actor network) and the Q-network (also known as the critic network). The policy network outputs a Gaussian distribution $\pi(*|s;\theta)$ to sample action for an input state $s$, where $\theta$ are the parameters of the network. Q-network outputs the evaluation value $Q(s,a;\Phi)$ for given part state $s$ and action $a$, where $\Phi$ are the parameters of the network. SAC uses an addition target Q-network to calculate target value for temporal difference (TD) update, whose parameters are denoted by $\Phi^\prime$.
A transition is denoted as a tuple ${\{s, a, R, s^\prime, d\}}$, where $R$ is the reward and $d$ indicates whether state $s^\prime$ is a terminated state.
All the transitions will be stored in a replay buffer $D$ and those two networks are trained by the data sampled from $D$.

The key change to the original SAC networks is that instead of outputting a single scale value to estimate the reward, we let the Q-network output a vector $(Q_g ,Q_c^0, \dots , Q_c^5) $ to estimate both $R_g$ and $R_c^i$. Accordingly, the reward $R$ in each experience is  a vector of $(R_g ,R_c^0, \dots , R_c^5) $. Note that  only $R_g$ is accumulated while $R_c^i$ is computed for each single step.
We found that this change made the training more stable and prevented the collision more effectively than directly combining all rewards together.


The loss function for training  the Q-network is then determined by temporal difference (TD) update:
\begin{equation}
    L_Q(\Phi)= [(Q_g(s,a;\Phi)-y_g(R_g,s^\prime,d))^2+
    \sum_{i=0}^5(Q_c^i(s,a;\Phi)-  \lambda R_c^i)^2)]
\end{equation}
with $\lambda =0.25$ balancing the two type of rewards and target value $y_g$ for $R_g$ defined as:
\begin{equation}
    y_g(R_g,s^\prime,d)=R_g+\gamma(1-d)\
    [Q_g(s^\prime,{\widetilde{a}}^\prime;\Phi^\prime)\
    -\alpha \log\pi(\widetilde{a}|s^\prime;\theta)],
\end{equation}
where ${\widetilde{a}}^\prime\sim\pi(*|s^\prime;\theta))$ is the sampled action. 
$\gamma = 0.99$ is the discount factor, and $\alpha $ is the temperature parameter that can be adjusted automatically to match an entropy target in expectation, to balance exploring the environment and maximizing reward.

For the policy network, in order to avoid self-collision of the gripper, i.e., the collision among different gripper components, we add a  self-collision loss  $L_{\text{self}}$ adapted from~\cite{liu2020deep} to the original loss:
\begin{equation}
	L(\theta) = L_{Q}(\theta) + \omega L_{\text{self}}(\theta),
\end{equation}
where $\omega = 100$ is the parameter to balance the two loss terms. The original loss function $L_{Q}(\theta)$ and the  self-collision loss $L_{\text{self}}$ for training  the policy network are defined as:
\begin{equation}
	L_Q(\theta)=[Q_g(s,\widetilde{a}(s;\theta))+(\sum_{i=0}^{5}Q_c^i(s,\widetilde{a}(s;\theta))-\alpha \log\pi({\widetilde{a}}|s;\theta)],
\end{equation}
\begin{equation}
	L_{\text{self}}(\theta) = \sum_{i=1}^{L} \sum_{j=1}^{N} \max (D(p_j(s,\widetilde{a}(s;\theta)), H_i(s,\widetilde{a}(s;\theta))), 0),
\end{equation}
where $\widetilde{a}\sim\pi(*|s;\theta))$ is the sampled action based on current state and network parameters, $L$ the number of gripper links, $N$ the number of points $p_j(s;\theta)$ sampled from each link, $H_i(s;\theta)$ the convex hull of each link after performing action $\widetilde{a}$ on state $s$, and $D$ the signed distance from a point to a convex hull. More details about the self-collision loss $L_{\text{self}}$ can be found in~\cite{liu2020deep}.  The reason why we add the self-collision loss directly for the policy network instead of defining a corresponding reward function is that the self-collision loss is differentiable and thus can be optimized directly.

\begin{figure}[!t]
	\centering
	\includegraphics[width=0.98\linewidth]{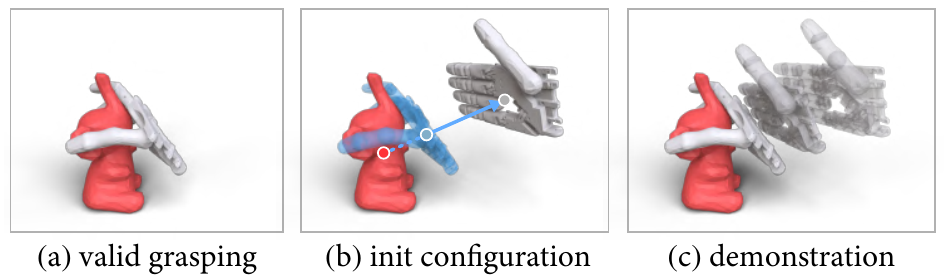}
	\caption{
		Method for demonstration generation. Given a valid grasping (a), we first push the gripper away from the object along the direction pointing from the object center to the palm center to reach a given distance, and then reset the gripper configuration with some random Gaussian noise to get the initial gripper configuration (b). The final demonstration (c) is generated by first moving the gripper close enough to the final position and then transiting to the final grasping configuration using linear interpolation.}
	\label{fig:demonstration}
\end{figure}

\paragraph{Training with demonstration.}
Note that the searching space of the action is extremely large and to make the training more efficient, we adopt the popular training with demonstration strategy.
To generate the demonstrations, as shown in Figure~\ref{fig:demonstration}, we first generate a valid grasp pose for the given object, and then move the gripper away from the object along the direction pointing from the object center to the palm center until the distance reaches $d = 20cm$, and then we reset the gripper configuration by setting all joint angles to be zero and add some random Gaussian noise  scaled based on the rotation limit of each joint to the gripper configuration to generate a random initial configuration. After getting the initial configuration of the gripper, to generate a motion sequence towards the corresponding final grasp pose and use it as the demonstration, we simply first move the gripper to the final position and then transit to the final grasping configuration using linear interpolation.

Note that the demonstration generated in this way is imperfect since the gripper may collide with the scene during the whole process, so unlike previous methods that usually use imitation learning for behavior cloning of the perfect demonstrations, we store the generated imperfect demonstration into the bootstrapping replay buffer and use reinforcement learning only. 

 In more detail, we use two replay buffers,  one for demonstration data with maximal size set to be $n_d = 5.0\times 10^4$ and the other for self-exploration data with maximal size set to be $n_s = 1.0\times 10^5$.
Before training, we always fill up the demonstration buffer and keep counting the total number $n_t$ of data generated in those two ways.
The probability of sampling data from the demonstration buffer is set to be $n_d/n_t$. Thus, more demonstration data will be sampled in the beginning to guide the network to learn a good initial policy quickly and hence speed up the training process.
Following~\cite{vecerik2017leveraging}, these samples are included in the update of both the actor and the critic.

\section{Results and Evaluation}

We first explain the experiment setup, the dataset we used, and then evaluate our method both qualitatively and quantitatively.


\subsection{Data preparation}
\rz{We first adopt the dataset provided by Liu et al.~\shortcite{liu2020deep}, which consists of 500 objects collected from four datasets.
We use the objects from  KIT Dataset \shortcite{kasper2012kit} and  GD Dataset \shortcite{kappler2015leveraging} as training data, and then test objects from YCB Dataset~\shortcite{calli2017yale}  and BigBIRD dataset~\shortcite{singh2014bigbird}.
We then further test our method on more objects from ContactPose Dataset \shortcite{brahmbhatt2020contactpose} and from 3DNet Dataset~\shortcite{wohlkinger20123dnet} when compared to baseline methods. 

To generate the demonstrations to guide the training of our network, we randomly select 100 grasp poses for each object in the training set from the pose set associated with that object, and synthesize imperfect reach-and-grasp motion as described in Section~\ref{sec:training}.
Note that since each object is placed on the table in our setting, we need to first filter out invalid grasp poses which collide with the table.



To generate our self-exploration data, we need to sample the initial configuration of the gripper.
For each object, we use its center to create a sphere with radius $r = 20 cm$, and sample points on the upper hemisphere as the origin of the local coordinate system of the gripper and rotate the gripper to make its palm face the object center \rh{and its thumb point upwards}.

As different initial configurations will lead to different reach-and-grasp results, to remove the bias of such initialization, we set a fixed set of initial configurations for each object to test.
The initialization details can be found in the supplementation material.
}

\begin{figure}[!t]
	\centering
	\includegraphics[width=0.98\linewidth]{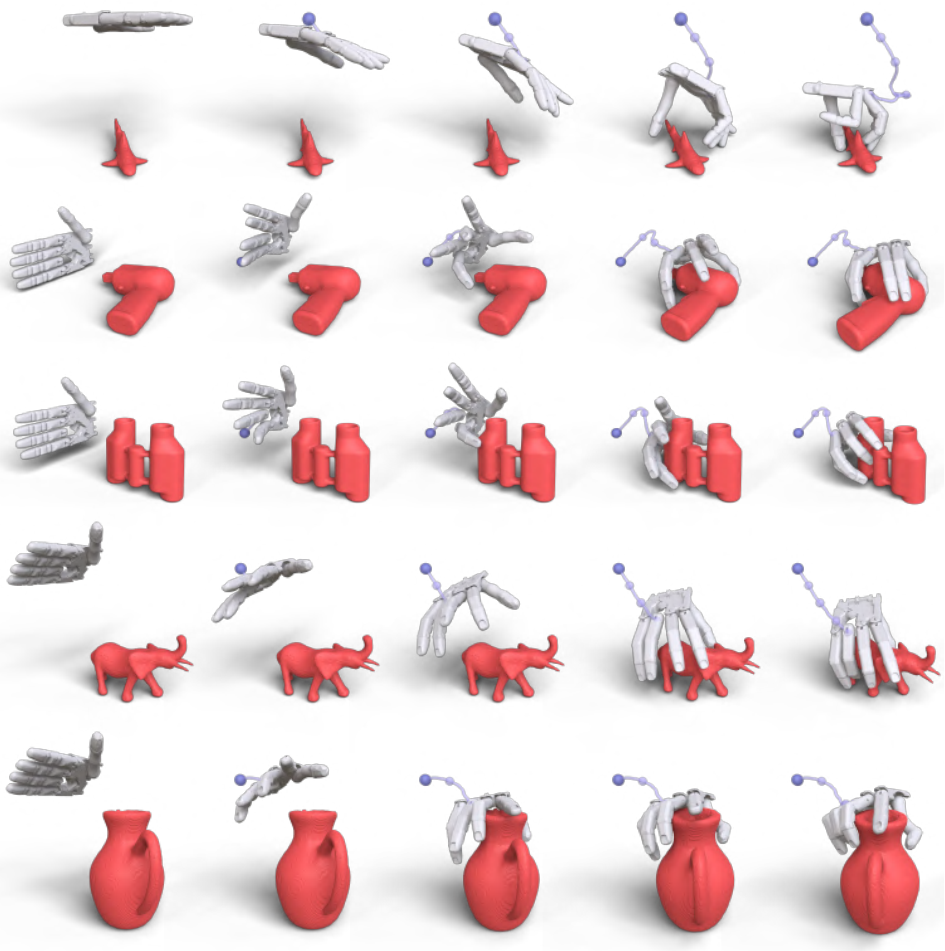}
	\caption{Gallery of results obtained with our method, where we show the input initial configuration of the gripper and four sampled frames during the approaching process with the final grasping pose on the right. \rz{The moving trajectory of the whole reach-and-grasp process is shown with the purple curve for each example.}
	}
	\label{fig:gallery}
\end{figure}

\subsection{Qualitative results}
\label{sec:qualitative}

\rz{Figure~\ref{fig:gallery} shows the gallery of results obtained with our method, where we show the initial input configuration of the gripper and four sampled frames during the approaching process with the final grasp pose on the right. Note that each example is shown with a view we selected to demonstrate the motion sequence more clearly, and we further add a purple curve to visualize moving trajectory together with sampled frames.
When inspecting these results, we see that the method is able to handle various shapes given different initial configurations of the gripper.
For example, for the small shark model shown in the first row, our method can adjust the gripper pose and joints precisely to pinch the model, while for the model shown in the second row, although the gripper starts from the position close to the table, our method is able to generate the moving sequence with the final grasp on the top of the model to avoid collision with the table. Our method is also able to deal with objects with more complex geometry. For example, to grasp the binoculars shown in the third row and the elephant shown in the fourth row, those four fingers tend to move together to form the gripping grasp with the thumb, while for the pitcher shown in the last row, fingers spread widely to better wrap its top.
}

\begin{figure}[!t]
	\centering
	\includegraphics[width=0.96\linewidth]{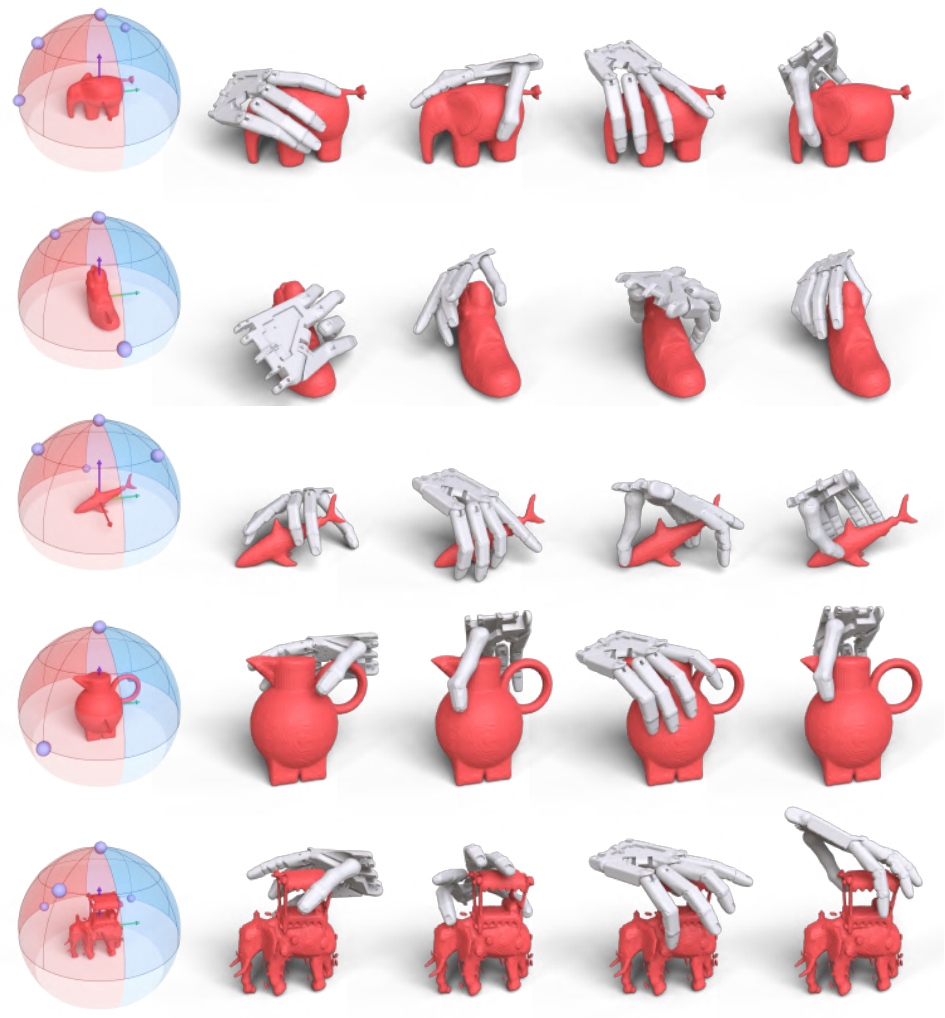}
	\caption{Example results where we fix the shape and plan the grasping with different initial gripper configurations. For each example, \rz{we show all four different initial configurations on the aligned hemisphere with purple dot on the left, and then their final grasping poses on the right.}
	}
	\label{fig:diff_init}
\end{figure}

Moreover, Figure~\ref{fig:diff_init} shows examples of results where we fix the shape and plan the grasping with different initial gripper configurations. \rz{For each example, we show four different initial configurations on the aligned hemisphere with purple dots on the left, and then their corresponding final grasp poses on the right.}
We see that various grasp poses can be generated even for the same shape when given different initial configurations. \rz{For example, for the elephant model shown in the first row, some final grasp poses tend to cover the head of the elephant while some others prefer wrapping the back of the model.
Similar results can also be observed for the shoe model shown in the second row.
For these models with far more complex geometry than the models we have in the training set, the final grasp poses obtained by our method can adapt well to their shapes while avoiding collision with the table at the same time.
Overall, our method is quite robust and can successfully plan grasping for different shapes with various geometries when given different initial gripper configurations.}

\begin{figure}[!t]
	\centering
	\includegraphics[width=0.98\linewidth]{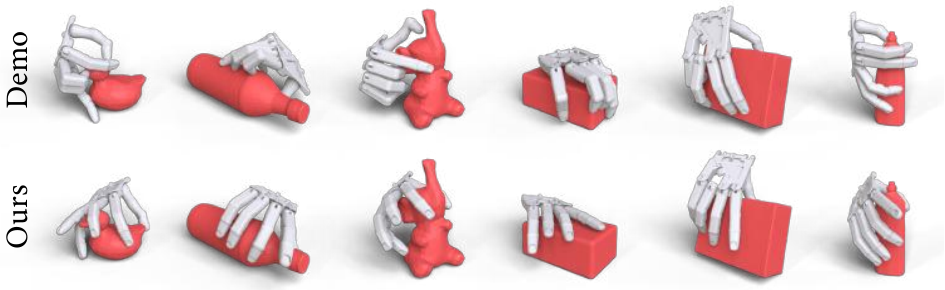}
	\caption{
		\rz{Visual comparisons of the final grasps generated by our method to the corresponding demonstration with the same object and same initial gripper configurations used for training. Note how our method is able to generate different and more natural grasp poses. }
	}
	\label{fig:train_demo}
\end{figure}

\rz{Note that all the objects tested in our experiments are unseen objects from different datasets, which shows that our method can generalize well to other objects instead of just remembering the grasp poses shown in the demonstrations. To further justify this, we compare the grasp pose synthesized by our method from the same initial configuration of the demonstration grasp in Figure~\ref{fig:train_demo}.
We can see that even starting from the same initial configuration, our method ends up with quite different and generally more natural grasp poses based on the learned policy.

}



\subsection{Quantitative evaluation}
\label{sec:quantitative}

\rz{To quantitatively evaluate our method, we first conduct ablation studies to justify several key design choices of our method, and then we provide comparisons to several baseline methods to show the superiority of our method.

\paragraph{Ablation studies.}
As explained in Section~\ref{sec:reward}, our goal is to learn a reach-and-grasp planner with high-quality final grasps as well as low penetration during the whole process, thus we use metrics to measure the final grasp and process penetration, respectively.
For the final grasp, we first compute the success rate $S$ among the whole testing set, where whether each final grasp is
successful or not is tested in the simulator. Then for all successful grasps, we compute the
average generalized $Q_1$~\cite{liu2020deep}
as a complementary metric to get a more detailed grasp quality measure.
For the process penetration, we further compute the percentage of the testing objects, of which the penetration of each frame during the whole reach-and-grasp process is smaller than a given threshold $\tau$, where $\tau = 1mm, 2mm$ and $3mm$.

Using IBS as the dynamic state representation is the key contribution of our method. Based on this, we further propose to use demonstrations, the vector-based representation of Q value, and a combination of the local and global encoder to boost the training and improve the performance.
To show the importance of all those design choices, we use the simplest version of our method as the baseline, denoted as ``IBS-G", which only uses a global encoder for IBS and a standard scalar Q value
and is trained without demonstrations, and then gradually add those key components one by one in the ablation studies to show how performance is boosted accordingly.
To further justify the superiority of IBS as the state representation, we also compare different versions of our method with either RGBD images or point clouds as state representation.
The evaluation results of the ablation studies  are shown in Table~\ref{tab:ablation_studies}.

%

\begin{table}[!t]%
	\caption{\rz{Ablation studies of our method.
	$S$  is the success rate of the final grasp and $Q_1$ is the mean generalized $Q_1$ value of  all successful grasps.
	The percentages of successful grasps whose process penetration is smaller than different thresholds $\tau = 1mm, 2mm$ and $3mm$ are also reported.
	More details about the metrics and the settings of different versions of our methods are provided in Section~\ref{sec:quantitative}.}}
	\label{tab:ablation_studies}
	\begin{minipage}{\columnwidth}
		\begin{center}
			\begin{tabular}{l|c|c|c|c|c}
				\hline
				\multicolumn{1}{l|}{\multirow{2}{*}{Method}} & \multicolumn{2}{c|}{Final grasp}  & \multicolumn{3}{c}{Process penetration}          \\ \cline{2-6}
			  &  $S$  & $Q_1$& $ \leq1mm$ & $\leq 2mm$ & $\leq 4mm$ \\ \hline\hline
			IBS-G & -   & - & - & - & - \\ \hline 
			+ Demo & 28.7$\%$ & 0.217 & 30.8$\%$ & 44.2$\%$ & 54.7$\%$ \\ \hline 
			+ Q-Vec & 65.8$\%$ & 0.215 & 47.2$\%$ & 63.3$\%$ & 79.1$\%$ \\ \hline
		   + IBS-L (Ours)& \textbf{68.3$\%$} & 0.207 & 58.3$\%$ & 75.7$\%$  & 89.7$\%$  \\ \hline\hline
			RGBD image & 9.1$\%$ & 0.173 & 59.6 $\%$ & 68.8$\%$ & 74.3$\%$ \\ \hline 
			Point cloud & 17.3$\%$ & 0.148 & 55.3$\%$ & 65.9$\%$ & 75.0$\%$ \\ \hline
			\end{tabular}
		\end{center}
	\end{minipage}
\end{table}%


%


\textit{Importance of demonstration.} Training with demonstrations is crucial to make our network be able to learn meaningful policy.
We can see in the Table~\ref{tab:ablation_studies} that without demonstration, the baseline ``IBS-G" cannot  learn any meaningful policy to perform valid grasping and thus the evaluation metrics cannot be reported, due to the high complexity of our problem and the corresponding large searching space.
When trained with demonstration, denoted as ``+ Demo'',
The network is able to learn a reasonable policy now, although the performance is not satisfactory enough with only $28.7\%$ success rate. Moreover, for the most of those successful grasps, the  penetration is higher than other settings.

\textit{Importance of  vector-based representation of Q value.}
Making the Q-Network output a vector instead of a single scale value provides better control of each individual gripper component.
To justify the benefit of such a design, we further enable this feature in our method and report the results in the third row of Table~\ref{tab:ablation_studies} , denoted as ``+ Q-Vec".
We see that compared to the results using a scalar Q value, the performance gets highly boosted, including both success rate and penetration avoidance.
The main reason is that when combining the grasp quality measure and the contacting measure together into one single value, it's hard for the network to learn which is the main reason to cause the change of the value, while using a vector-based representation can provide a more clear guidance for the network to learn.

\textit{Importance of multi-level encoder.} To further show the importance of the multi-level encoder used for feature extraction of IBS, i.e., using both global and local encoders and then concatenating the features together, 
we further add the local encoder to get the full version of our method, and the result is shown in the fourth row of Table~\ref{tab:ablation_studies} , denoted as ``+ IBS-L".
We can see that the performance is better than other settings, which shows that the component-based IBS partition and the corresponding local encoder can help get more information and give better control of those gripper components.

\textit{Importance of IBS as state representation.} The introduction of IBS to represent the intermediate state during the whole planning process and encode the interaction between the gripper and the scene is one of our key contributions. To show the importance of our IBS encoding, we compare our method to two alternative ways of interaction representation, i.e., using either RGBD images or point clouds of the scene and the gripper, while all the other design choices are the same to our method, including using demonstrations and vector-based representation of Q value.

For RGBD image presentation, we use visual information captured by a simulated hand-mounted camera in its upright-oriented pose that translates together with the hand but does not rotate. The visual input consists of an RGBD image and the corresponding segmentation information, where each pixel can belong to the object, the gripper or the background.
We then replace the encoder with similar architecture to that of  the network used in~\cite{Jain-ICRA-19}.
More details about this baseline can be found in the supplementary material.

For point cloud representation, \rz{we use the scene point cloud and gripper point cloud directly as input.
	For the features to be encoded, we remove the features that contain interaction information from the list of features we defined for IBS points in Section~\ref{sec:state} and keep all the remaining ones, which include the coordinate $c_s \in R^3$ and foreground/background  indicator $b_s \in {0,1}$ for each scene point and  coordinate $c_g \in R^3$, gripper component indicate $c_g \in  \{0, 1\}^6$ and gripper side indicator $a_g \in [-1, 1]$ for each gripper point.
The encoder is similar to that we used for IBS-G.
More details about this baseline can also be found in the supplementary material. }

The corresponding performances are shown in the last two rows of  Table~\ref{tab:ablation_studies}, denoted as ``RGBD image" and ``Point cloud", respectively.  Note that our method with either setting performs poorly .
We think the main reason is that they cannot provide enough spatial information between the object and the gripper when the demonstrations provided are imperfect and with large variations in initial configurations and final grasp poses, while the IBS  we used provides a more informative representation.

}

\rz{\paragraph{Comparison to baselines.}
Strategies for the reach-and-grasp task can be roughly divided into two categories, one synthesizes the grasp first and then plans the transfer from the initial configuration to the final grasp pose and the other optimizes the whole process directly.
To give more detailed comparisons to previous methods,  we organize the baseline methods according to those two different types of strategies and analyze the results separately.

\begin{table}[!t]%
		\caption{
			\rz{Quantitative comparison to two-step baseline methods with ``grasp synthesis + motion planning”. where final grasp poses are synthesized using different methods, including Liu\shortcite{liu2020deep}, GraspIt!~\cite{miller2004graspit}, and our method, and motions from the same gripper configuration to those final grasp poses are planned using the same method. We conduct the comparison on the YCB dataset, and report the success rates of the final grasp, motion planning, and overall execution in the simulator, where ``Avg" refers to the average success rate among all testing samples and `` Top-1" refers to the average success rate of all testing object. One test sample means one object with one initial configuration.}
	}
	\label{tab:comp_mp}
	\begin{minipage}{\columnwidth}
		\begin{center}
			\begin{tabular}{l|l|l|l|l|l|l}
				\hline
				\multicolumn{1}{l|}{\multirow{2}{*}{Method}} & \multicolumn{2}{c|}{Final grasp}                            & \multicolumn{2}{c|}{Motion Plan}                             & \multicolumn{2}{c}{Overall}      \\ \cline{2-7}
				\multicolumn{1}{l|}{}                        & \multicolumn{1}{l|}{Avg} & \multicolumn{1}{l|}{Top-1} & \multicolumn{1}{l|}{Avg} & \multicolumn{1}{l|}{Top-1} & \multicolumn{1}{l|}{Avg} & Top-1 \\ \hline\hline
			Liu\shortcite{liu2020deep}& 13.6\%  & 24.0\% & 34.4\% & 44.0\%  & 8.0\% & 12.0\%  \\ \hline
			GraspIt! & 52.1\%  & 62.0\% & 36.3\% & 54.0\%  & 22.6\% & 42.0\% \\ \hline
			Ours & \textbf{68.3\%} & \textbf{100.0\%} & \textbf{58.3\%} & \textbf{100.0\%}  & \textbf{43.2\%} & \textbf{90.0\%} \\ \hline
			\end{tabular}
		\end{center}
	\end{minipage}
\end{table}%

For the first type of baselines, we compare the final grasps obtained using our method to those synthesized using the method in~\cite{liu2020deep} and GraspIt!~\cite{miller2004graspit}.
For the final grasp pose generated via GraspIt!, we select the one closest to the initial gripper configuration from a pre-sampled set of candidate grasp poses based on the distance metric proposed in~\cite{di2008posedis}
To further generate the whole reach-and-grasp process, we use the same motion planning method~\cite{kavraki1996PRM} from The Open Motion Planning Library (OMPL)~\cite{sucan2012open} to plan a collision-free  moving trajectory from the initial configuration to the global pose of the final pose.
We can execute the whole process in our dynamics simulator to see if the object can be successfully grasped in the end to get an overall success.
In more detail, once the gripper in the rest pose reaches its final position and orientation following the planned trajectory, we start using the interface provided by the simulator to move fingers towards the target joint states unless it contacts with the scene.
When all fingers stop moving, we use the simulator to execute the grasping test.
\rh{Accordingly, we can compute the success rate for different stages. 
The first one is ``final grasp'', where we test whether the gripper with the given final configuration can successfully grasp and lift the object in the dynamics simulator.
The second one is ``motion plan'', where we check whether the given planner can find a feasible trajectory to transit the gripper from the initial configuration to the final configuration. 
The third one is ``overall'', where we consider the whole process to be successful if and only if both  final grasp and motion plan succeed.
}

Table~\ref{tab:comp_mp} reports the success rate of the final grasp, motion plan, and the overall process executed in the dynamic simulator .
Note that the success rates are computed for different settings.
``Avg" refers to the average success rate among all testing samples, while `` Top-1" refers to the average success rate of all testing objects, where each object is tested with a fixed set of initial configurations and is considered to be successfully grasped if any of  those initial configurations leads to a successful grasp.
We can see that using the final grasp obtained via our method gets consistently better results.

\begin{figure}[!t]
	\centering
	\includegraphics[width=0.96\linewidth]{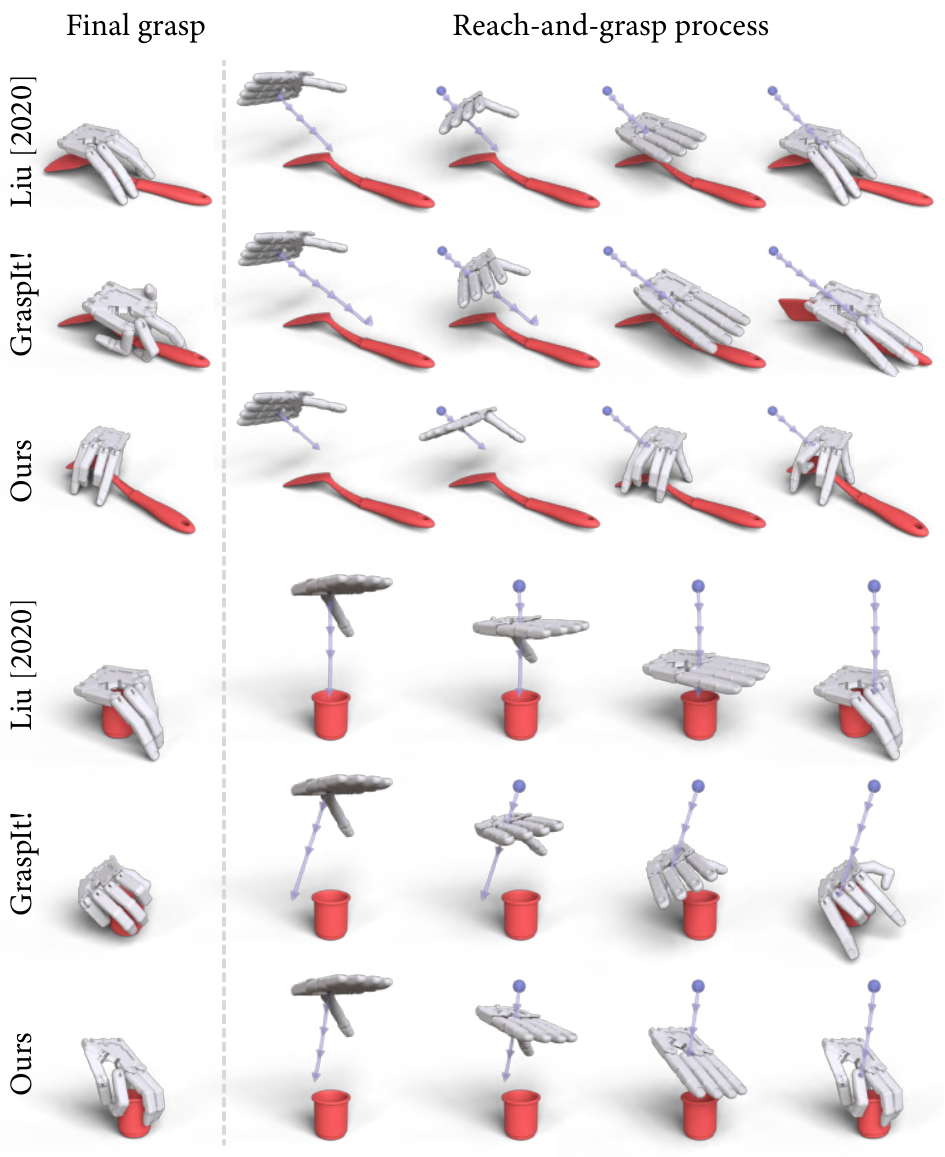}
	\caption{ \rz{Visual comparisons to two-step baseline method with ``grasp synthesis + motion planning” . For each example, we show the final grasp synthesized by each method on the left and several key frames of the reach-and-grasp process with the offline planned trajectory in purple on the right. 
		}
	}
	\label{fig:comp_mp}
\end{figure}

\rz{Figure~\ref{fig:comp_mp} shows some visual comparisons of those methods.  
For each example, we show the final grasp synthesized by each method on the left and several key frames of the reach-and-grasp process with the offline planned trajectory in purple on the right.
From the results, the method of Liu et al.~\shortcite{liu2020deep} cannot handle the tabletop well: It tends to generate a grasp pose with finger touching the table and does not form a valid grasp pose, especially for flat-shaped objects shown in the first row.
Although GraspIt! can generate much more successful grasps, it usually fails to plan a collision-free moving path for the gripper from the given initial configuration to the final pose, and even with successfully planned path, the gripper configuration is more likely to be changed during the reaching process to avoid collision with the tabletop, which leads to a lower overall success rate. Note how the final grasp pose after the reaching process is different from the planned final poses shown in the fifth row.
But for the final grasp pose generated by our method, we have taken the tabletop into the consideration during the whole reaching process, thus on the one hand, our method gets a much higher success rate of the final grasp, one the other hand, even using the same motion planning method instead of the own motion planned by our method, we can still get the best overall success rate.

}

\begin{table}[!t]%
	\caption{\rz{Quantitative comparison to the primitive-based method (PBM) on four datasets. We report the overall success rates of execution in the simulator, where ``Avg" refers to the average success rate among all testing samples and `` Top-1" refers to the average success rate of all testing object. One test sample means one object with one initial configuration.} }
	\label{tab:dataset}
	\begin{minipage}{\columnwidth}
		\begin{center}
			\resizebox{\textwidth}{!}{
			\begin{tabular}{l|l|l|l|l|l|l|l|l}
				\hline
				\multicolumn{1}{l|}{\multirow{2}{*}{Method}} & \multicolumn{2}{c|}{YCB*}  & \multicolumn{2}{c|}{BIGBIRD*} & \multicolumn{2}{c|}{ContactPose} & \multicolumn{2}{c}{3DNet*} \\ \cline{2-9}
				\multicolumn{1}{l|}{} & \multicolumn{1}{l|}{Avg} & \multicolumn{1}{l|}{Top-1} & \multicolumn{1}{l|}{Avg} & \multicolumn{1}{l|}{Top-1} & \multicolumn{1}{l|}{Avg} & \multicolumn{1}{l|}{Top-1} & \multicolumn{1}{l|}{Avg} & \multicolumn{1}{l}{Top-1} \\ \hline\hline
				PBM & 57.0\% & 92.0\% & 47.2\% & 89.0\% & 44.5\% & 84.0\% & 48.7\% & 90.0\% \\ \hline
				Ours & \textbf{58.6\%} & \textbf{100.0\%} & \textbf{47.7\%} & \textbf{95.6\%} & \textbf{48.0\%} & \textbf{88.0\%} & \textbf{52.3\%} & \textbf{100.0\%} \\ \hline
			\end{tabular}
			}
		\end{center}
	\end{minipage}
\end{table}%

\begin{figure}[!t]
	\centering
	\includegraphics[width=0.96\linewidth]{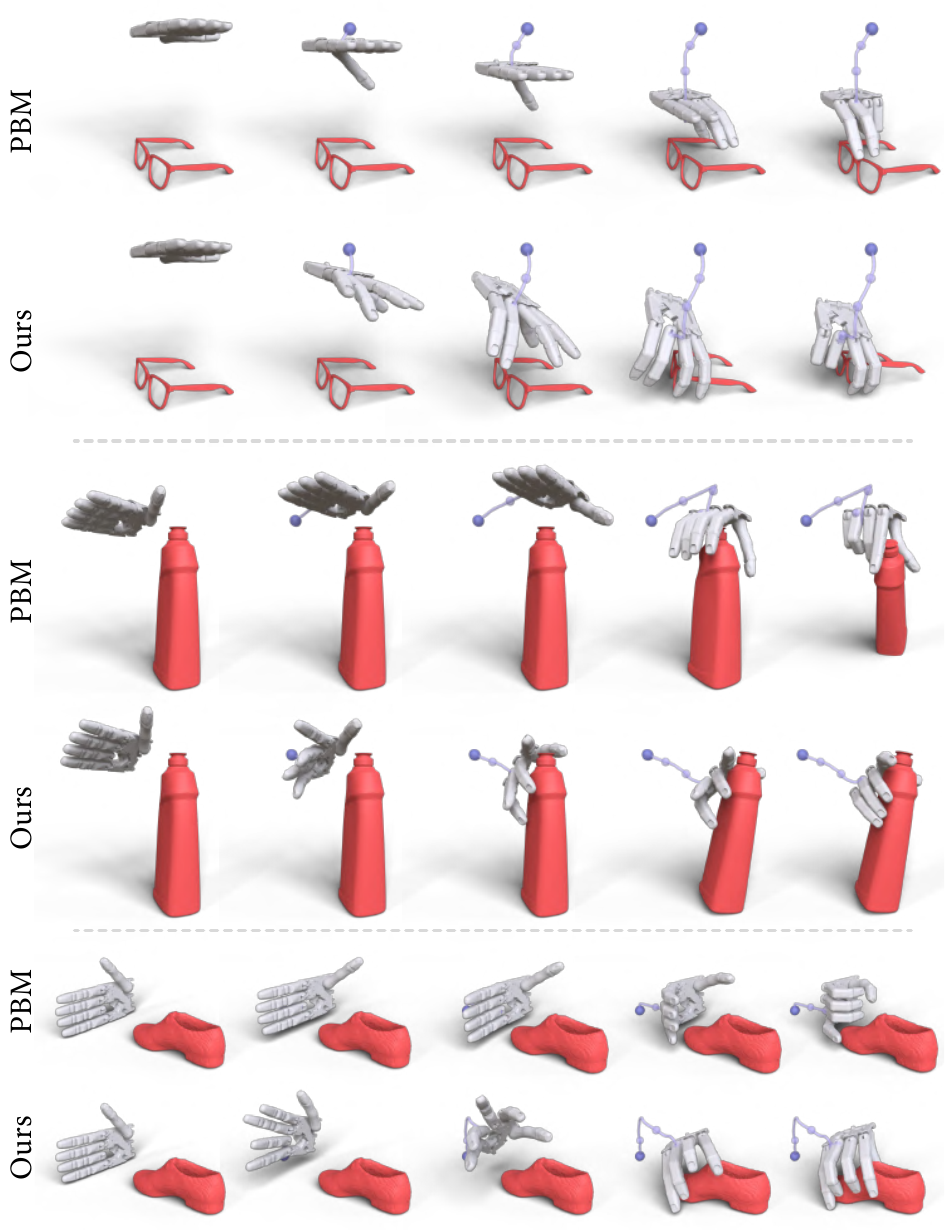}
	\caption{ \rz{Visual comparisons to the primitive-based method (PBM). 
	For each example, we show the initial configuration of the gripper on the left, final grasping pose on the right, and three sampled frames during the reaching process in the middle. The purple curve indicates the moving trajectory before the current frame. 
	From top to bottom, the primitives used in these examples for PBM are ``Pinch", ``Top",  and ``Lateral", respectively.}
	}
	\label{fig:comp_pbm}
\end{figure}


For the second type of baselines, we compare the whole reach-and-grasp process obtained via our method to a heuristic primitive-based grasping method.
Inspired by the work of \cite{della2019learning}, we adopt three grasping primitives, including ``Pinch", ``Top", and ``Lateral", and for each testing initial configuration, we select the closest primitive based on the geodesic distance on the sphere to execute the corresponding grasp.
Note that  ``Top" and ``Pinch" primitives have the same start gripper configurations, and we choose the one that achieves higher performance to get the final result.
More details about this primitive-based method (PBM) can be found in the supplementary material.
}

\rz{Table~\ref{tab:dataset} reports the comparisons of ``Avg" and `` Top-1"  success rate of overall execution in the simulator on four different datasets. We can see that our method outperforms the primitive-based method on all the datasets.
The main reason is the reaching path of each primitive grasp is fixed and cannot adapt well to the various geometry of different objects, while our method can keep targeting the object with the guidance of the encoded IBS.
Figure~\ref{fig:comp_pbm} shows some visual comparisons between our method and PBM.
Note how the final grasp poses of the primitive-based method deviate from the object due to the shape complexity.
For example, neither the final grasps of PBM shown in the first and third examples touches the object.

One interesting aspect of our method we would like to highlight is that when executing the reach-and-grasp process in the simulator,  the pose of the target object may change due to collision with the gripper, and our method is still able to dynamically adapt to its new pose, follow the moving object and grasp it successfully. See the grasp result of the bottle shown in the fourth row of Figure~\ref{fig:comp_pbm}.
Other than the adaption to dynamic changes of the target object, our method is also robust to partial observations.
Examples can be found in the supplementary materials.
}

\section{Discussion and Future Work}

\begin{figure}[!t]
	\centering
	\includegraphics[width=0.96\linewidth]{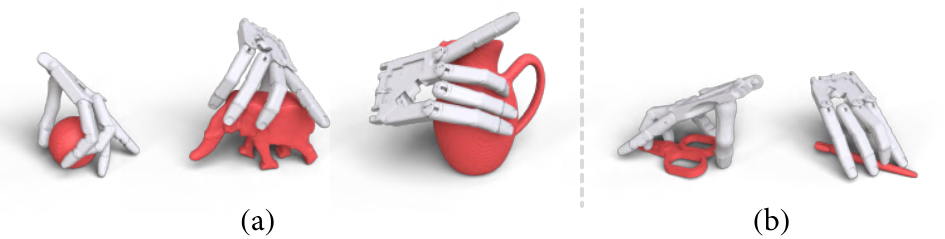}
	\caption{ \rz{Failure cases.  (a) Unnatural grasping poses. (b) Unsuccessful grasps for flat-shaped target objects.} }
	\label{fig:failure}
\end{figure}

We have presented an RL-based method to jointly learn grasp and motion planning for high-DOF grippers.
We advocate the use of Interaction Bisector Surface to characterize the fine-grained spatial relationship between the gripper and the target object. We found that IBS is surprisingly effective as a state/observation representation of deep RL since it well informs the fine-grained control of each finger with spatial relation against the target object. Together with a few critical designs of the learning model and strategy, our method learns high-quality grasping with smooth reaching motion.

\rz{
Our method has the following limitations:
\begin{itemize}
	\item Our RL model adopts success signal in execution as well as Q1 metric for measuring grasp quality and does not explicitly define the naturalness of a grasp. Therefore, there is the case that a generated grasp is successful but looks unnatural. For example, some successful grasps could have an unbent finger that looks implausible; see Figure~\ref{fig:failure}(a).
    \item The other reward we use is devised to avoid collision during the reach-and-grasp process. This makes our method unable to learn picking up a flat shape lying on the table. See Figure~\ref{fig:failure}(b) for an example.
\end{itemize}
}





%
%

\rz{As future work, we would like to conduct further investigation on the following four aspects:
\begin{itemize}
    \item To further improve the performance of our method, we can try more complicated feature encoding other than the current PointNet and MPLs, and further, take more dynamic information of each frame such as velocity into account.
    \item To generate more natural grasps, it is necessary to investigate grasp quality metrics that can better reflect grasp naturalness. This could be learned from human grasping datasets such as ContactPose~\cite{brahmbhatt2020contactpose} .
    \item To make our method be able to grasp flat-shaped objects, we can relax the collision constraint and even utilize the collision with the environment to help lift the object and achieve a successful grasp as in~\cite{eppner2015exploitation} .
  \item To be able to conduct real robot implementation, we need to study how to perform sim-to-real policy transfer, overcoming the domain gap between simulated observation and real visual perception, as well as the gap between simplified static environment and real dynamic scenes.
\end{itemize}
}

\section{ACKNOWLEDGEMENTS}
{
This work was supported in parts by
NSFC (61872250, 62132021, U2001206, U21B2023, 62161146005),
GD Talent Plan (2019JC05X328), 
GD Natural Science Foundation (2021B1515020085),
DEGP Key Project (2018KZDXM058, 2020SFKC059),
National Key R\&D Program of China (2018AAA0102200), 
Shenzhen Science and Technology Program (RCYX20210609103121030, RCJC20200714114435012, JCYJ20210324120213036), 
and Guangdong Laboratory of Artificial Intelligence and Digital Economy (SZ).}

\bibliographystyle{ACM-Reference-Format}
\bibliography{Grasping_arxiv}


\begin{thebibliography}{73}


\ifx \showCODEN    \undefined \def \showCODEN     #1{\unskip}     \fi
\ifx \showDOI      \undefined \def \showDOI       #1{#1}\fi
\ifx \showISBNx    \undefined \def \showISBNx     #1{\unskip}     \fi
\ifx \showISBNxiii \undefined \def \showISBNxiii  #1{\unskip}     \fi
\ifx \showISSN     \undefined \def \showISSN      #1{\unskip}     \fi
\ifx \showLCCN     \undefined \def \showLCCN      #1{\unskip}     \fi
\ifx \shownote     \undefined \def \shownote      #1{#1}          \fi
\ifx \showarticletitle \undefined \def \showarticletitle #1{#1}   \fi
\ifx \showURL      \undefined \def \showURL       {\relax}        \fi
\providecommand\bibfield[2]{#2}
\providecommand\bibinfo[2]{#2}
\providecommand\natexlab[1]{#1}
\providecommand\showeprint[2][]{arXiv:#2}

\bibitem[\protect\citeauthoryear{Aleotti and Caselli}{Aleotti and
  Caselli}{2012}]%
        {aleotti20123d}
\bibfield{author}{\bibinfo{person}{Jacopo Aleotti} {and}
  \bibinfo{person}{Stefano Caselli}.} \bibinfo{year}{2012}\natexlab{}.
\newblock \showarticletitle{A 3D shape segmentation approach for robot grasping
  by parts}.
\newblock \bibinfo{journal}{\emph{Robotics and Autonomous Systems}}
  \bibinfo{volume}{60}, \bibinfo{number}{3} (\bibinfo{year}{2012}),
  \bibinfo{pages}{358--366}.
\newblock


\bibitem[\protect\citeauthoryear{Andrychowicz, Baker, Chociej, Jozefowicz,
  McGrew, Pachocki, Petron, Plappert, Powell, Ray, et~al\mbox{.}}{Andrychowicz
  et~al\mbox{.}}{2020}]%
        {andrychowicz2020learning}
\bibfield{author}{\bibinfo{person}{OpenAI:~Marcin Andrychowicz},
  \bibinfo{person}{Bowen Baker}, \bibinfo{person}{Maciek Chociej},
  \bibinfo{person}{Rafal Jozefowicz}, \bibinfo{person}{Bob McGrew},
  \bibinfo{person}{Jakub Pachocki}, \bibinfo{person}{Arthur Petron},
  \bibinfo{person}{Matthias Plappert}, \bibinfo{person}{Glenn Powell},
  \bibinfo{person}{Alex Ray}, {et~al\mbox{.}}} \bibinfo{year}{2020}\natexlab{}.
\newblock \showarticletitle{Learning dexterous in-hand manipulation}.
\newblock \bibinfo{journal}{\emph{The International Journal of Robotics
  Research}} \bibinfo{volume}{39}, \bibinfo{number}{1} (\bibinfo{year}{2020}),
  \bibinfo{pages}{3--20}.
\newblock


\bibitem[\protect\citeauthoryear{Bohg, Morales, Asfour, and Kragic}{Bohg
  et~al\mbox{.}}{2013}]%
        {bohg2013data}
\bibfield{author}{\bibinfo{person}{Jeannette Bohg}, \bibinfo{person}{Antonio
  Morales}, \bibinfo{person}{Tamim Asfour}, {and} \bibinfo{person}{Danica
  Kragic}.} \bibinfo{year}{2013}\natexlab{}.
\newblock \showarticletitle{Data-driven grasp synthesis—a survey}.
\newblock \bibinfo{journal}{\emph{IEEE Transactions on Robotics}}
  \bibinfo{volume}{30}, \bibinfo{number}{2} (\bibinfo{year}{2013}),
  \bibinfo{pages}{289--309}.
\newblock


\bibitem[\protect\citeauthoryear{Brahmbhatt, Tang, Twigg, Kemp, and
  Hays}{Brahmbhatt et~al\mbox{.}}{2020}]%
        {brahmbhatt2020contactpose}
\bibfield{author}{\bibinfo{person}{Samarth Brahmbhatt},
  \bibinfo{person}{Chengcheng Tang}, \bibinfo{person}{Christopher~D Twigg},
  \bibinfo{person}{Charles~C Kemp}, {and} \bibinfo{person}{James Hays}.}
  \bibinfo{year}{2020}\natexlab{}.
\newblock \showarticletitle{ContactPose: A dataset of grasps with object
  contact and hand pose}. In \bibinfo{booktitle}{\emph{Computer Vision--ECCV
  2020: 16th European Conference, Glasgow, UK, August 23--28, 2020,
  Proceedings, Part XIII 16}}. Springer, \bibinfo{pages}{361--378}.
\newblock


\bibitem[\protect\citeauthoryear{Calli, Singh, Bruce, Walsman, Konolige,
  Srinivasa, Abbeel, and Dollar}{Calli et~al\mbox{.}}{2017}]%
        {calli2017yale}
\bibfield{author}{\bibinfo{person}{Berk Calli}, \bibinfo{person}{Arjun Singh},
  \bibinfo{person}{James Bruce}, \bibinfo{person}{Aaron Walsman},
  \bibinfo{person}{Kurt Konolige}, \bibinfo{person}{Siddhartha Srinivasa},
  \bibinfo{person}{Pieter Abbeel}, {and} \bibinfo{person}{Aaron~M Dollar}.}
  \bibinfo{year}{2017}\natexlab{}.
\newblock \showarticletitle{Yale-CMU-Berkeley dataset for robotic manipulation
  research}.
\newblock \bibinfo{journal}{\emph{The International Journal of Robotics
  Research}} \bibinfo{volume}{36}, \bibinfo{number}{3} (\bibinfo{year}{2017}),
  \bibinfo{pages}{261--268}.
\newblock


\bibitem[\protect\citeauthoryear{Chen and Burdick}{Chen and Burdick}{1993}]%
        {chen1993finding}
\bibfield{author}{\bibinfo{person}{I-Ming Chen} {and} \bibinfo{person}{Joel~W
  Burdick}.} \bibinfo{year}{1993}\natexlab{}.
\newblock \showarticletitle{Finding antipodal point grasps on irregularly
  shaped objects}.
\newblock \bibinfo{journal}{\emph{IEEE transactions on Robotics and
  Automation}} \bibinfo{volume}{9}, \bibinfo{number}{4} (\bibinfo{year}{1993}),
  \bibinfo{pages}{507--512}.
\newblock


\bibitem[\protect\citeauthoryear{Collet and Srinivasa}{Collet and
  Srinivasa}{2010}]%
        {collet2010efficient}
\bibfield{author}{\bibinfo{person}{Alvaro Collet} {and}
  \bibinfo{person}{Siddhartha~S Srinivasa}.} \bibinfo{year}{2010}\natexlab{}.
\newblock \showarticletitle{Efficient multi-view object recognition and full
  pose estimation}. In \bibinfo{booktitle}{\emph{2010 IEEE International
  Conference on Robotics and Automation}}. IEEE, \bibinfo{pages}{2050--2055}.
\newblock


\bibitem[\protect\citeauthoryear{Della~Santina, Arapi, Averta, Damiani, Fiore,
  Settimi, Catalano, Bacciu, Bicchi, and Bianchi}{Della~Santina
  et~al\mbox{.}}{2019}]%
        {della2019learning}
\bibfield{author}{\bibinfo{person}{Cosimo Della~Santina},
  \bibinfo{person}{Visar Arapi}, \bibinfo{person}{Giuseppe Averta},
  \bibinfo{person}{Francesca Damiani}, \bibinfo{person}{Gaia Fiore},
  \bibinfo{person}{Alessandro Settimi}, \bibinfo{person}{Manuel~G Catalano},
  \bibinfo{person}{Davide Bacciu}, \bibinfo{person}{Antonio Bicchi}, {and}
  \bibinfo{person}{Matteo Bianchi}.} \bibinfo{year}{2019}\natexlab{}.
\newblock \showarticletitle{Learning from humans how to grasp: a data-driven
  architecture for autonomous grasping with anthropomorphic soft hands}.
\newblock \bibinfo{journal}{\emph{IEEE Robotics and Automation Letters}}
  \bibinfo{volume}{4}, \bibinfo{number}{2} (\bibinfo{year}{2019}),
  \bibinfo{pages}{1533--1540}.
\newblock


\bibitem[\protect\citeauthoryear{Depierre, Dellandr{\'e}a, and Chen}{Depierre
  et~al\mbox{.}}{2018}]%
        {depierre2018jacquard}
\bibfield{author}{\bibinfo{person}{Amaury Depierre}, \bibinfo{person}{Emmanuel
  Dellandr{\'e}a}, {and} \bibinfo{person}{Liming Chen}.}
  \bibinfo{year}{2018}\natexlab{}.
\newblock \showarticletitle{Jacquard: A large scale dataset for robotic grasp
  detection}. In \bibinfo{booktitle}{\emph{2018 IEEE/RSJ International
  Conference on Intelligent Robots and Systems (IROS)}}. IEEE,
  \bibinfo{pages}{3511--3516}.
\newblock


\bibitem[\protect\citeauthoryear{Di~Gregorio}{Di~Gregorio}{2008}]%
        {di2008posedis}
\bibfield{author}{\bibinfo{person}{Raffaele Di~Gregorio}.}
  \bibinfo{year}{2008}\natexlab{}.
\newblock \showarticletitle{A novel point of view to define the distance
  between two rigid-body poses}.
\newblock In \bibinfo{booktitle}{\emph{Advances in robot kinematics: Analysis
  and design}}. \bibinfo{publisher}{Springer}, \bibinfo{pages}{361--369}.
\newblock


\bibitem[\protect\citeauthoryear{Eppner, Deimel, Alvarez-Ruiz, Maertens, and
  Brock}{Eppner et~al\mbox{.}}{2015}]%
        {eppner2015exploitation}
\bibfield{author}{\bibinfo{person}{Clemens Eppner}, \bibinfo{person}{Raphael
  Deimel}, \bibinfo{person}{Jos{\'e} Alvarez-Ruiz}, \bibinfo{person}{Marianne
  Maertens}, {and} \bibinfo{person}{Oliver Brock}.}
  \bibinfo{year}{2015}\natexlab{}.
\newblock \showarticletitle{Exploitation of environmental constraints in human
  and robotic grasping}.
\newblock \bibinfo{journal}{\emph{The International Journal of Robotics
  Research}} \bibinfo{volume}{34}, \bibinfo{number}{7} (\bibinfo{year}{2015}),
  \bibinfo{pages}{1021--1038}.
\newblock


\bibitem[\protect\citeauthoryear{Fang, Bai, Hinterstoisser, Savarese, and
  Kalakrishnan}{Fang et~al\mbox{.}}{2018}]%
        {fang2018mtda}
\bibfield{author}{\bibinfo{person}{Kuan Fang}, \bibinfo{person}{Yunfei Bai},
  \bibinfo{person}{Stefan Hinterstoisser}, \bibinfo{person}{Silvio Savarese},
  {and} \bibinfo{person}{Mrinal Kalakrishnan}.}
  \bibinfo{year}{2018}\natexlab{}.
\newblock \showarticletitle{Multi-Task Domain Adaptation for Deep Learning of
  Instance Grasping from Simulation}.
\newblock \bibinfo{journal}{\emph{IEEE International Conference on Robotics and
  Automation (ICRA)}} (\bibinfo{year}{2018}).
\newblock


\bibitem[\protect\citeauthoryear{Ferrari and Canny}{Ferrari and Canny}{1992}]%
        {ferrari1992planning}
\bibfield{author}{\bibinfo{person}{Carlo Ferrari} {and} \bibinfo{person}{John~F
  Canny}.} \bibinfo{year}{1992}\natexlab{}.
\newblock \showarticletitle{Planning optimal grasps.}. In
  \bibinfo{booktitle}{\emph{ICRA}}, Vol.~\bibinfo{volume}{3}.
  \bibinfo{pages}{2290--2295}.
\newblock


\bibitem[\protect\citeauthoryear{Ficuciello, Migliozzi, Laudante, Falco, and
  Siciliano}{Ficuciello et~al\mbox{.}}{2019}]%
        {ficuciello2019vision}
\bibfield{author}{\bibinfo{person}{F Ficuciello}, \bibinfo{person}{A
  Migliozzi}, \bibinfo{person}{G Laudante}, \bibinfo{person}{P Falco}, {and}
  \bibinfo{person}{B Siciliano}.} \bibinfo{year}{2019}\natexlab{}.
\newblock \showarticletitle{Vision-based grasp learning of an anthropomorphic
  hand-arm system in a synergy-based control framework}.
\newblock \bibinfo{journal}{\emph{Science robotics}} \bibinfo{volume}{4},
  \bibinfo{number}{26} (\bibinfo{year}{2019}).
\newblock


\bibitem[\protect\citeauthoryear{Ficuciello, Zaccara, and Siciliano}{Ficuciello
  et~al\mbox{.}}{2016}]%
        {ficuciello2016synergy}
\bibfield{author}{\bibinfo{person}{Fanny Ficuciello}, \bibinfo{person}{Damiano
  Zaccara}, {and} \bibinfo{person}{Bruno Siciliano}.}
  \bibinfo{year}{2016}\natexlab{}.
\newblock \showarticletitle{Synergy-based policy improvement with path
  integrals for anthropomorphic hands}. In \bibinfo{booktitle}{\emph{2016
  IEEE/RSJ International Conference on Intelligent Robots and Systems (IROS)}}.
  IEEE, \bibinfo{pages}{1940--1945}.
\newblock


\bibitem[\protect\citeauthoryear{Fontanals, Dang-Vu, Porges, Rosell, and
  Roa}{Fontanals et~al\mbox{.}}{2014}]%
        {fontanals2014integrated}
\bibfield{author}{\bibinfo{person}{Joan Fontanals}, \bibinfo{person}{Bao-Anh
  Dang-Vu}, \bibinfo{person}{Oliver Porges}, \bibinfo{person}{Jan Rosell},
  {and} \bibinfo{person}{M{\'a}ximo~A Roa}.} \bibinfo{year}{2014}\natexlab{}.
\newblock \showarticletitle{Integrated grasp and motion planning using
  independent contact regions}. In \bibinfo{booktitle}{\emph{2014 IEEE-RAS
  International Conference on Humanoid Robots}}. IEEE,
  \bibinfo{pages}{887--893}.
\newblock


\bibitem[\protect\citeauthoryear{Gualtieri, Ten~Pas, Saenko, and
  Platt}{Gualtieri et~al\mbox{.}}{2016}]%
        {gualtieri2016high}
\bibfield{author}{\bibinfo{person}{Marcus Gualtieri}, \bibinfo{person}{Andreas
  Ten~Pas}, \bibinfo{person}{Kate Saenko}, {and} \bibinfo{person}{Robert
  Platt}.} \bibinfo{year}{2016}\natexlab{}.
\newblock \showarticletitle{High precision grasp pose detection in dense
  clutter}. In \bibinfo{booktitle}{\emph{2016 IEEE/RSJ International Conference
  on Intelligent Robots and Systems (IROS)}}. IEEE, \bibinfo{pages}{598--605}.
\newblock


\bibitem[\protect\citeauthoryear{Haarnoja, Zhou, Hartikainen, Tucker, Ha, Tan,
  Kumar, Zhu, Gupta, Abbeel, et~al\mbox{.}}{Haarnoja et~al\mbox{.}}{2018}]%
        {haarnoja2018soft}
\bibfield{author}{\bibinfo{person}{Tuomas Haarnoja}, \bibinfo{person}{Aurick
  Zhou}, \bibinfo{person}{Kristian Hartikainen}, \bibinfo{person}{George
  Tucker}, \bibinfo{person}{Sehoon Ha}, \bibinfo{person}{Jie Tan},
  \bibinfo{person}{Vikash Kumar}, \bibinfo{person}{Henry Zhu},
  \bibinfo{person}{Abhishek Gupta}, \bibinfo{person}{Pieter Abbeel},
  {et~al\mbox{.}}} \bibinfo{year}{2018}\natexlab{}.
\newblock \showarticletitle{Soft actor-critic algorithms and applications}.
\newblock \bibinfo{journal}{\emph{arXiv preprint arXiv:1812.05905}}
  (\bibinfo{year}{2018}).
\newblock


\bibitem[\protect\citeauthoryear{Hang, Stork, Pollard, and Kragic}{Hang
  et~al\mbox{.}}{2017}]%
        {7812687}
\bibfield{author}{\bibinfo{person}{Kaiyu Hang}, \bibinfo{person}{Johannes~A.
  Stork}, \bibinfo{person}{Nancy~S. Pollard}, {and} \bibinfo{person}{Danica
  Kragic}.} \bibinfo{year}{2017}\natexlab{}.
\newblock \showarticletitle{A Framework for Optimal Grasp Contact Planning}.
\newblock \bibinfo{journal}{\emph{IEEE Robotics and Automation Letters}}
  \bibinfo{volume}{2}, \bibinfo{number}{2} (\bibinfo{year}{2017}),
  \bibinfo{pages}{704--711}.
\newblock
\urldef\tempurl%
\url{https://doi.org/10.1109/LRA.2017.2651381}
\showDOI{\tempurl}


\bibitem[\protect\citeauthoryear{Hu, Zhu, van Kaick, Liu, Shamir, and Zhang}{Hu
  et~al\mbox{.}}{2015}]%
        {hu2015interaction}
\bibfield{author}{\bibinfo{person}{Ruizhen Hu}, \bibinfo{person}{Chenyang Zhu},
  \bibinfo{person}{Oliver van Kaick}, \bibinfo{person}{Ligang Liu},
  \bibinfo{person}{Ariel Shamir}, {and} \bibinfo{person}{Hao Zhang}.}
  \bibinfo{year}{2015}\natexlab{}.
\newblock \showarticletitle{Interaction context (ICON) towards a geometric
  functionality descriptor}.
\newblock \bibinfo{journal}{\emph{ACM Transactions on Graphics}}
  \bibinfo{volume}{34}, \bibinfo{number}{4} (\bibinfo{year}{2015}),
  \bibinfo{pages}{1--12}.
\newblock


\bibitem[\protect\citeauthoryear{Jain, Li, Singhal, Rajeswaran, Kumar, and
  Todorov}{Jain et~al\mbox{.}}{2019}]%
        {Jain-ICRA-19}
\bibfield{author}{\bibinfo{person}{Divye Jain}, \bibinfo{person}{Andrew Li},
  \bibinfo{person}{Shivam Singhal}, \bibinfo{person}{Aravind Rajeswaran},
  \bibinfo{person}{Vikash Kumar}, {and} \bibinfo{person}{Emanuel Todorov}.}
  \bibinfo{year}{2019}\natexlab{}.
\newblock \showarticletitle{{Learning Deep Visuomotor Policies for Dexterous
  Hand Manipulation}}. In \bibinfo{booktitle}{\emph{International Conference on
  Robotics and Automation (ICRA)}}.
\newblock


\bibitem[\protect\citeauthoryear{James, Wohlhart, Kalakrishnan, Kalashnikov,
  Irpan, Ibarz, Levine, Hadsell, and Bousmalis}{James et~al\mbox{.}}{2019}]%
        {james2019sim}
\bibfield{author}{\bibinfo{person}{Stephen James}, \bibinfo{person}{Paul
  Wohlhart}, \bibinfo{person}{Mrinal Kalakrishnan}, \bibinfo{person}{Dmitry
  Kalashnikov}, \bibinfo{person}{Alex Irpan}, \bibinfo{person}{Julian Ibarz},
  \bibinfo{person}{Sergey Levine}, \bibinfo{person}{Raia Hadsell}, {and}
  \bibinfo{person}{Konstantinos Bousmalis}.} \bibinfo{year}{2019}\natexlab{}.
\newblock \showarticletitle{Sim-to-real via sim-to-sim: Data-efficient robotic
  grasping via randomized-to-canonical adaptation networks}. In
  \bibinfo{booktitle}{\emph{Proc. IEEE Conf. on Computer Vision \& Pattern
  Recognition}}. \bibinfo{pages}{12627--12637}.
\newblock


\bibitem[\protect\citeauthoryear{Kalashnikov, Irpan, Pastor, Ibarz, Herzog,
  Jang, Quillen, Holly, Kalakrishnan, Vanhoucke, et~al\mbox{.}}{Kalashnikov
  et~al\mbox{.}}{2018}]%
        {kalashnikov2018qt}
\bibfield{author}{\bibinfo{person}{Dmitry Kalashnikov}, \bibinfo{person}{Alex
  Irpan}, \bibinfo{person}{Peter Pastor}, \bibinfo{person}{Julian Ibarz},
  \bibinfo{person}{Alexander Herzog}, \bibinfo{person}{Eric Jang},
  \bibinfo{person}{Deirdre Quillen}, \bibinfo{person}{Ethan Holly},
  \bibinfo{person}{Mrinal Kalakrishnan}, \bibinfo{person}{Vincent Vanhoucke},
  {et~al\mbox{.}}} \bibinfo{year}{2018}\natexlab{}.
\newblock \showarticletitle{Qt-opt: Scalable deep reinforcement learning for
  vision-based robotic manipulation}.
\newblock \bibinfo{journal}{\emph{arXiv preprint arXiv:1806.10293}}
  (\bibinfo{year}{2018}).
\newblock


\bibitem[\protect\citeauthoryear{Kappler, Bohg, and Schaal}{Kappler
  et~al\mbox{.}}{2015}]%
        {kappler2015leveraging}
\bibfield{author}{\bibinfo{person}{Daniel Kappler}, \bibinfo{person}{Jeannette
  Bohg}, {and} \bibinfo{person}{Stefan Schaal}.}
  \bibinfo{year}{2015}\natexlab{}.
\newblock \showarticletitle{Leveraging big data for grasp planning}. In
  \bibinfo{booktitle}{\emph{2015 IEEE International Conference on Robotics and
  Automation (ICRA)}}. IEEE, \bibinfo{pages}{4304--4311}.
\newblock


\bibitem[\protect\citeauthoryear{Karunratanakul, Yang, Zhang, Black, Muandet,
  and Tang}{Karunratanakul et~al\mbox{.}}{2020}]%
        {karunratanakul2020grasping}
\bibfield{author}{\bibinfo{person}{Korrawe Karunratanakul},
  \bibinfo{person}{Jinlong Yang}, \bibinfo{person}{Yan Zhang},
  \bibinfo{person}{Michael Black}, \bibinfo{person}{Krikamol Muandet}, {and}
  \bibinfo{person}{Siyu Tang}.} \bibinfo{year}{2020}\natexlab{}.
\newblock \showarticletitle{Grasping Field: Learning Implicit Representations
  for Human Grasps}.
\newblock \bibinfo{journal}{\emph{arXiv preprint arXiv:2008.04451}}
  (\bibinfo{year}{2020}).
\newblock


\bibitem[\protect\citeauthoryear{Kasper, Xue, and Dillmann}{Kasper
  et~al\mbox{.}}{2012}]%
        {kasper2012kit}
\bibfield{author}{\bibinfo{person}{Alexander Kasper}, \bibinfo{person}{Zhixing
  Xue}, {and} \bibinfo{person}{R{\"u}diger Dillmann}.}
  \bibinfo{year}{2012}\natexlab{}.
\newblock \showarticletitle{The KIT object models database: An object model
  database for object recognition, localization and manipulation in service
  robotics}.
\newblock \bibinfo{journal}{\emph{The International Journal of Robotics
  Research}} \bibinfo{volume}{31}, \bibinfo{number}{8} (\bibinfo{year}{2012}),
  \bibinfo{pages}{927--934}.
\newblock


\bibitem[\protect\citeauthoryear{Kavraki, Svestka, Latombe, and
  Overmars}{Kavraki et~al\mbox{.}}{1996}]%
        {kavraki1996PRM}
\bibfield{author}{\bibinfo{person}{Lydia~E Kavraki}, \bibinfo{person}{Petr
  Svestka}, \bibinfo{person}{J-C Latombe}, {and} \bibinfo{person}{Mark~H
  Overmars}.} \bibinfo{year}{1996}\natexlab{}.
\newblock \showarticletitle{Probabilistic roadmaps for path planning in
  high-dimensional configuration spaces}.
\newblock \bibinfo{journal}{\emph{IEEE transactions on Robotics and
  Automation}} \bibinfo{volume}{12}, \bibinfo{number}{4}
  (\bibinfo{year}{1996}), \bibinfo{pages}{566--580}.
\newblock


\bibitem[\protect\citeauthoryear{Kiatos and Malassiotis}{Kiatos and
  Malassiotis}{2019}]%
        {kiatos2019grasping}
\bibfield{author}{\bibinfo{person}{Marios Kiatos} {and}
  \bibinfo{person}{Sotiris Malassiotis}.} \bibinfo{year}{2019}\natexlab{}.
\newblock \showarticletitle{Grasping Unknown Objects by Exploiting
  Complementarity with Robot Hand Geometry}. In
  \bibinfo{booktitle}{\emph{International Conference on Computer Vision
  Systems}}. Springer, \bibinfo{pages}{88--97}.
\newblock


\bibitem[\protect\citeauthoryear{Kim, Won, Cho, Kim, Lee, Bhak, and Kim}{Kim
  et~al\mbox{.}}{2006}]%
        {kim2006interaction}
\bibfield{author}{\bibinfo{person}{Chong-Min Kim}, \bibinfo{person}{Chung-In
  Won}, \bibinfo{person}{Youngsong Cho}, \bibinfo{person}{Donguk Kim},
  \bibinfo{person}{Sunghoon Lee}, \bibinfo{person}{Jonghwa Bhak}, {and}
  \bibinfo{person}{Deok-Soo Kim}.} \bibinfo{year}{2006}\natexlab{}.
\newblock \showarticletitle{Interaction interfaces in proteins via the Voronoi
  diagram of atoms}.
\newblock \bibinfo{journal}{\emph{Computer-Aided Design}} \bibinfo{volume}{38},
  \bibinfo{number}{11} (\bibinfo{year}{2006}), \bibinfo{pages}{1192--1204}.
\newblock


\bibitem[\protect\citeauthoryear{Kleeberger, Bormann, Kraus, and
  Huber}{Kleeberger et~al\mbox{.}}{2020}]%
        {kleeberger2020survey}
\bibfield{author}{\bibinfo{person}{Kilian Kleeberger}, \bibinfo{person}{Richard
  Bormann}, \bibinfo{person}{Werner Kraus}, {and} \bibinfo{person}{Marco~F
  Huber}.} \bibinfo{year}{2020}\natexlab{}.
\newblock \showarticletitle{A survey on learning-based robotic grasping}.
\newblock \bibinfo{journal}{\emph{Current Robotics Reports}}
  (\bibinfo{year}{2020}), \bibinfo{pages}{1--11}.
\newblock


\bibitem[\protect\citeauthoryear{Levine, Pastor, Krizhevsky, Ibarz, and
  Quillen}{Levine et~al\mbox{.}}{2018}]%
        {levine2018learning}
\bibfield{author}{\bibinfo{person}{Sergey Levine}, \bibinfo{person}{Peter
  Pastor}, \bibinfo{person}{Alex Krizhevsky}, \bibinfo{person}{Julian Ibarz},
  {and} \bibinfo{person}{Deirdre Quillen}.} \bibinfo{year}{2018}\natexlab{}.
\newblock \showarticletitle{Learning hand-eye coordination for robotic grasping
  with deep learning and large-scale data collection}.
\newblock \bibinfo{journal}{\emph{The International Journal of Robotics
  Research}} \bibinfo{volume}{37}, \bibinfo{number}{4-5}
  (\bibinfo{year}{2018}), \bibinfo{pages}{421--436}.
\newblock


\bibitem[\protect\citeauthoryear{Liu}{Liu}{2009}]%
        {liu2009dextrous}
\bibfield{author}{\bibinfo{person}{C~Karen Liu}.}
  \bibinfo{year}{2009}\natexlab{}.
\newblock \showarticletitle{Dextrous manipulation from a grasping pose}.
\newblock \bibinfo{journal}{\emph{ACM Trans. on Graphics (Proc. SIGGRAPH)}}
  \bibinfo{volume}{28}, \bibinfo{number}{3} (\bibinfo{year}{2009}),
  \bibinfo{pages}{59:1--59:6}.
\newblock


\bibitem[\protect\citeauthoryear{Liu, Pan, Xu, Ganguly, and Manocha}{Liu
  et~al\mbox{.}}{2019}]%
        {liu2019grasp}
\bibfield{author}{\bibinfo{person}{Min Liu}, \bibinfo{person}{Zherong Pan},
  \bibinfo{person}{Kai Xu}, \bibinfo{person}{Kanishka Ganguly}, {and}
  \bibinfo{person}{Dinesh Manocha}.} \bibinfo{year}{2019}\natexlab{}.
\newblock \showarticletitle{Generating Grasp Poses for A High-DOF Gripper Using
  Neural Networks}. In \bibinfo{booktitle}{\emph{Proceedings of the IEEE/RSJ
  International Conference on Intelligent Robots and Systems}}.
  \bibinfo{pages}{1518--1525}.
\newblock


\bibitem[\protect\citeauthoryear{Liu, Pan, Xu, Ganguly, and Manocha}{Liu
  et~al\mbox{.}}{2020b}]%
        {liu2020deep}
\bibfield{author}{\bibinfo{person}{Min Liu}, \bibinfo{person}{Zherong Pan},
  \bibinfo{person}{Kai Xu}, \bibinfo{person}{Kanishka Ganguly}, {and}
  \bibinfo{person}{Dinesh Manocha}.} \bibinfo{year}{2020}\natexlab{b}.
\newblock \showarticletitle{Deep Differentiable Grasp Planner for High-DOF
  Grippers}. In \bibinfo{booktitle}{\emph{Proceedings of the Robotics: Science
  and Systems 2020}}.
\newblock


\bibitem[\protect\citeauthoryear{Liu, Pan, Xu, and Manocha}{Liu
  et~al\mbox{.}}{2020a}]%
        {9120282}
\bibfield{author}{\bibinfo{person}{Min Liu}, \bibinfo{person}{Zherong Pan},
  \bibinfo{person}{Kai Xu}, {and} \bibinfo{person}{Dinesh Manocha}.}
  \bibinfo{year}{2020}\natexlab{a}.
\newblock \showarticletitle{New Formulation of Mixed-Integer Conic Programming
  for Globally Optimal Grasp Planning}.
\newblock \bibinfo{journal}{\emph{IEEE Robotics and Automation Letters}}
  \bibinfo{volume}{5}, \bibinfo{number}{3} (\bibinfo{year}{2020}),
  \bibinfo{pages}{4663--4670}.
\newblock
\urldef\tempurl%
\url{https://doi.org/10.1109/LRA.2020.3003280}
\showDOI{\tempurl}


\bibitem[\protect\citeauthoryear{Lu, Chenna, Sundaralingam, and Hermans}{Lu
  et~al\mbox{.}}{2020a}]%
        {lu2020planning}
\bibfield{author}{\bibinfo{person}{Qingkai Lu}, \bibinfo{person}{Kautilya
  Chenna}, \bibinfo{person}{Balakumar Sundaralingam}, {and}
  \bibinfo{person}{Tucker Hermans}.} \bibinfo{year}{2020}\natexlab{a}.
\newblock \showarticletitle{Planning multi-fingered grasps as probabilistic
  inference in a learned deep network}.
\newblock In \bibinfo{booktitle}{\emph{Robotics Research}}.
  \bibinfo{publisher}{Springer}, \bibinfo{pages}{455--472}.
\newblock


\bibitem[\protect\citeauthoryear{Lu, Van~der Merwe, Sundaralingam, and
  Hermans}{Lu et~al\mbox{.}}{2020b}]%
        {lu2020multifingered}
\bibfield{author}{\bibinfo{person}{Qingkai Lu}, \bibinfo{person}{Mark Van~der
  Merwe}, \bibinfo{person}{Balakumar Sundaralingam}, {and}
  \bibinfo{person}{Tucker Hermans}.} \bibinfo{year}{2020}\natexlab{b}.
\newblock \showarticletitle{Multifingered Grasp Planning via Inference in Deep
  Neural Networks: Outperforming Sampling by Learning Differentiable Models}.
\newblock \bibinfo{journal}{\emph{IEEE Robotics \& Automation Magazine}}
  (\bibinfo{year}{2020}).
\newblock


\bibitem[\protect\citeauthoryear{Mahler, Liang, Niyaz, Laskey, Doan, Liu,
  Aparicio, and Goldberg}{Mahler et~al\mbox{.}}{2017}]%
        {inproceedingsDexNetTwo}
\bibfield{author}{\bibinfo{person}{Jeffrey Mahler}, \bibinfo{person}{Jacky
  Liang}, \bibinfo{person}{Sherdil Niyaz}, \bibinfo{person}{Michael Laskey},
  \bibinfo{person}{Richard Doan}, \bibinfo{person}{Xinyu Liu},
  \bibinfo{person}{Juan Aparicio}, {and} \bibinfo{person}{Ken Goldberg}.}
  \bibinfo{year}{2017}\natexlab{}.
\newblock \showarticletitle{Dex-Net 2.0: Deep Learning to Plan Robust Grasps
  with Synthetic Point Clouds and Analytic Grasp Metrics}. In
  \bibinfo{booktitle}{\emph{Proceedings of Robotics: Science and Systems}}.
\newblock
\urldef\tempurl%
\url{https://doi.org/10.15607/RSS.2017.XIII.058}
\showDOI{\tempurl}


\bibitem[\protect\citeauthoryear{Mahler, Matl, Liu, Li, Gealy, and
  Goldberg}{Mahler et~al\mbox{.}}{2018}]%
        {mahler2018dex}
\bibfield{author}{\bibinfo{person}{Jeffrey Mahler}, \bibinfo{person}{Matthew
  Matl}, \bibinfo{person}{Xinyu Liu}, \bibinfo{person}{Albert Li},
  \bibinfo{person}{David Gealy}, {and} \bibinfo{person}{Ken Goldberg}.}
  \bibinfo{year}{2018}\natexlab{}.
\newblock \showarticletitle{Dex-net 3.0: Computing robust vacuum suction grasp
  targets in point clouds using a new analytic model and deep learning}. In
  \bibinfo{booktitle}{\emph{2018 IEEE International Conference on robotics and
  automation (ICRA)}}. IEEE, \bibinfo{pages}{5620--5627}.
\newblock


\bibitem[\protect\citeauthoryear{Mahler, Pokorny, Hou, Roderick, Laskey, Aubry,
  Kohlhoff, Kr{\"o}ger, Kuffner, and Goldberg}{Mahler et~al\mbox{.}}{2016}]%
        {mahler2016dex}
\bibfield{author}{\bibinfo{person}{Jeffrey Mahler}, \bibinfo{person}{Florian~T
  Pokorny}, \bibinfo{person}{Brian Hou}, \bibinfo{person}{Melrose Roderick},
  \bibinfo{person}{Michael Laskey}, \bibinfo{person}{Mathieu Aubry},
  \bibinfo{person}{Kai Kohlhoff}, \bibinfo{person}{Torsten Kr{\"o}ger},
  \bibinfo{person}{James Kuffner}, {and} \bibinfo{person}{Ken Goldberg}.}
  \bibinfo{year}{2016}\natexlab{}.
\newblock \showarticletitle{Dex-net 1.0: A cloud-based network of 3d objects
  for robust grasp planning using a multi-armed bandit model with correlated
  rewards}. In \bibinfo{booktitle}{\emph{2016 IEEE international conference on
  robotics and automation (ICRA)}}. IEEE, \bibinfo{pages}{1957--1964}.
\newblock


\bibitem[\protect\citeauthoryear{Maldonado, Klank, and Beetz}{Maldonado
  et~al\mbox{.}}{2010}]%
        {maldonado2010robotic}
\bibfield{author}{\bibinfo{person}{Alexis Maldonado}, \bibinfo{person}{Ulrich
  Klank}, {and} \bibinfo{person}{Michael Beetz}.}
  \bibinfo{year}{2010}\natexlab{}.
\newblock \showarticletitle{Robotic grasping of unmodeled objects using
  time-of-flight range data and finger torque information}. In
  \bibinfo{booktitle}{\emph{International Conference on Intelligent Robots and
  Systems}}. IEEE, \bibinfo{pages}{2586--2591}.
\newblock


\bibitem[\protect\citeauthoryear{Mandikal and Grauman}{Mandikal and
  Grauman}{2021}]%
        {mandikal2020dexterous}
\bibfield{author}{\bibinfo{person}{Priyanka Mandikal} {and}
  \bibinfo{person}{Kristen Grauman}.} \bibinfo{year}{2021}\natexlab{}.
\newblock \showarticletitle{Dexterous robotic grasping with object-centric
  visual affordances}. In \bibinfo{booktitle}{\emph{2021 IEEE international
  conference on robotics and automation (ICRA)}}. IEEE.
\newblock


\bibitem[\protect\citeauthoryear{Miller and Allen}{Miller and Allen}{2000}]%
        {miller2000graspit}
\bibfield{author}{\bibinfo{person}{Andrew~T Miller} {and}
  \bibinfo{person}{Peter~K Allen}.} \bibinfo{year}{2000}\natexlab{}.
\newblock \showarticletitle{Graspit!: A versatile simulator for grasp
  analysis}. In \bibinfo{booktitle}{\emph{in Proc. of the ASME Dynamic Systems
  and Control Division}}. Citeseer.
\newblock


\bibitem[\protect\citeauthoryear{Miller and Allen}{Miller and Allen}{2004}]%
        {miller2004graspit}
\bibfield{author}{\bibinfo{person}{Andrew~T Miller} {and}
  \bibinfo{person}{Peter~K Allen}.} \bibinfo{year}{2004}\natexlab{}.
\newblock \showarticletitle{Graspit! a versatile simulator for robotic
  grasping}.
\newblock \bibinfo{journal}{\emph{IEEE Robotics \& Automation Magazine}}
  \bibinfo{volume}{11}, \bibinfo{number}{4} (\bibinfo{year}{2004}),
  \bibinfo{pages}{110--122}.
\newblock


\bibitem[\protect\citeauthoryear{Mnih, Kavukcuoglu, Silver, Rusu, Veness,
  Bellemare, Graves, Riedmiller, Fidjeland, Ostrovski, et~al\mbox{.}}{Mnih
  et~al\mbox{.}}{2015}]%
        {mnih2015human}
\bibfield{author}{\bibinfo{person}{Volodymyr Mnih}, \bibinfo{person}{Koray
  Kavukcuoglu}, \bibinfo{person}{David Silver}, \bibinfo{person}{Andrei~A
  Rusu}, \bibinfo{person}{Joel Veness}, \bibinfo{person}{Marc~G Bellemare},
  \bibinfo{person}{Alex Graves}, \bibinfo{person}{Martin Riedmiller},
  \bibinfo{person}{Andreas~K Fidjeland}, \bibinfo{person}{Georg Ostrovski},
  {et~al\mbox{.}}} \bibinfo{year}{2015}\natexlab{}.
\newblock \showarticletitle{Human-level control through deep reinforcement
  learning}.
\newblock \bibinfo{journal}{\emph{nature}} \bibinfo{volume}{518},
  \bibinfo{number}{7540} (\bibinfo{year}{2015}), \bibinfo{pages}{529--533}.
\newblock


\bibitem[\protect\citeauthoryear{Monforte, Ficuciello, and Siciliano}{Monforte
  et~al\mbox{.}}{2019}]%
        {monforte2019multifunctional}
\bibfield{author}{\bibinfo{person}{Marco Monforte}, \bibinfo{person}{Fanny
  Ficuciello}, {and} \bibinfo{person}{Bruno Siciliano}.}
  \bibinfo{year}{2019}\natexlab{}.
\newblock \showarticletitle{Multifunctional principal component analysis for
  human-like grasping}.
\newblock In \bibinfo{booktitle}{\emph{Human Friendly Robotics}}.
  \bibinfo{publisher}{Springer}, \bibinfo{pages}{47--58}.
\newblock


\bibitem[\protect\citeauthoryear{Morrison, Corke, and Leitner}{Morrison
  et~al\mbox{.}}{2018}]%
        {morrison2018closing}
\bibfield{author}{\bibinfo{person}{Douglas Morrison}, \bibinfo{person}{Peter
  Corke}, {and} \bibinfo{person}{J{\"u}rgen Leitner}.}
  \bibinfo{year}{2018}\natexlab{}.
\newblock \showarticletitle{Closing the loop for robotic grasping: A real-time,
  generative grasp synthesis approach}.
\newblock \bibinfo{journal}{\emph{arXiv preprint arXiv:1804.05172}}
  (\bibinfo{year}{2018}).
\newblock


\bibitem[\protect\citeauthoryear{Nguyen}{Nguyen}{1988}]%
        {nguyen1988constructing}
\bibfield{author}{\bibinfo{person}{Van-Duc Nguyen}.}
  \bibinfo{year}{1988}\natexlab{}.
\newblock \showarticletitle{Constructing force-closure grasps}.
\newblock \bibinfo{journal}{\emph{The International Journal of Robotics
  Research}} \bibinfo{volume}{7}, \bibinfo{number}{3} (\bibinfo{year}{1988}),
  \bibinfo{pages}{3--16}.
\newblock


\bibitem[\protect\citeauthoryear{Pan, Zhang, Tu, and Gao}{Pan
  et~al\mbox{.}}{2022}]%
        {9626457}
\bibfield{author}{\bibinfo{person}{Zherong Pan}, \bibinfo{person}{Duo Zhang},
  \bibinfo{person}{Changhe Tu}, {and} \bibinfo{person}{Xifeng Gao}.}
  \bibinfo{year}{2022}\natexlab{}.
\newblock \showarticletitle{Planning of Power Grasps Using Infinite Program
  Under Complementary Constraints}.
\newblock \bibinfo{journal}{\emph{IEEE Robotics and Automation Letters}}
  \bibinfo{volume}{7}, \bibinfo{number}{1} (\bibinfo{year}{2022}),
  \bibinfo{pages}{650--657}.
\newblock


\bibitem[\protect\citeauthoryear{Peng, Andrychowicz, Zaremba, and Abbeel}{Peng
  et~al\mbox{.}}{2018}]%
        {peng2018sim}
\bibfield{author}{\bibinfo{person}{Xue~Bin Peng}, \bibinfo{person}{Marcin
  Andrychowicz}, \bibinfo{person}{Wojciech Zaremba}, {and}
  \bibinfo{person}{Pieter Abbeel}.} \bibinfo{year}{2018}\natexlab{}.
\newblock \showarticletitle{Sim-to-real transfer of robotic control with
  dynamics randomization}. In \bibinfo{booktitle}{\emph{2018 IEEE international
  conference on robotics and automation (ICRA)}}. IEEE,
  \bibinfo{pages}{3803--3810}.
\newblock


\bibitem[\protect\citeauthoryear{Pirk, Krs, Hu, Rajasekaran, Kang, Yoshiyasu,
  Benes, and Guibas}{Pirk et~al\mbox{.}}{2017}]%
        {pirk2017understanding}
\bibfield{author}{\bibinfo{person}{S{\"o}ren Pirk}, \bibinfo{person}{Vojtech
  Krs}, \bibinfo{person}{Kaimo Hu}, \bibinfo{person}{Suren~Deepak Rajasekaran},
  \bibinfo{person}{Hao Kang}, \bibinfo{person}{Yusuke Yoshiyasu},
  \bibinfo{person}{Bedrich Benes}, {and} \bibinfo{person}{Leonidas~J Guibas}.}
  \bibinfo{year}{2017}\natexlab{}.
\newblock \showarticletitle{Understanding and exploiting object interaction
  landscapes}.
\newblock \bibinfo{journal}{\emph{ACM Transactions on Graphics}}
  \bibinfo{volume}{36}, \bibinfo{number}{3} (\bibinfo{year}{2017}),
  \bibinfo{pages}{1--14}.
\newblock


\bibitem[\protect\citeauthoryear{Pollard and Zordan}{Pollard and
  Zordan}{2005}]%
        {pollard2005physically}
\bibfield{author}{\bibinfo{person}{Nancy~S Pollard} {and}
  \bibinfo{person}{Victor~Brian Zordan}.} \bibinfo{year}{2005}\natexlab{}.
\newblock \showarticletitle{Physically based grasping control from example}. In
  \bibinfo{booktitle}{\emph{Proceedings of the 2005 ACM SIGGRAPH/Eurographics
  symposium on Computer animation}}. \bibinfo{pages}{311--318}.
\newblock


\bibitem[\protect\citeauthoryear{Qi, Su, Mo, and Guibas}{Qi
  et~al\mbox{.}}{2017}]%
        {qi2017pointnet}
\bibfield{author}{\bibinfo{person}{Charles~R Qi}, \bibinfo{person}{Hao Su},
  \bibinfo{person}{Kaichun Mo}, {and} \bibinfo{person}{Leonidas~J Guibas}.}
  \bibinfo{year}{2017}\natexlab{}.
\newblock \showarticletitle{Pointnet: Deep learning on point sets for 3d
  classification and segmentation}. In \bibinfo{booktitle}{\emph{Proc. IEEE
  Conf. on Computer Vision \& Pattern Recognition}}. \bibinfo{pages}{652--660}.
\newblock


\bibitem[\protect\citeauthoryear{Quillen, Jang, Nachum, Finn, Ibarz, and
  Levine}{Quillen et~al\mbox{.}}{2018}]%
        {quillen2018deep}
\bibfield{author}{\bibinfo{person}{Deirdre Quillen}, \bibinfo{person}{Eric
  Jang}, \bibinfo{person}{Ofir Nachum}, \bibinfo{person}{Chelsea Finn},
  \bibinfo{person}{Julian Ibarz}, {and} \bibinfo{person}{Sergey Levine}.}
  \bibinfo{year}{2018}\natexlab{}.
\newblock \showarticletitle{Deep reinforcement learning for vision-based
  robotic grasping: A simulated comparative evaluation of off-policy methods}.
  In \bibinfo{booktitle}{\emph{2018 IEEE International Conference on Robotics
  and Automation (ICRA)}}. IEEE, \bibinfo{pages}{6284--6291}.
\newblock


\bibitem[\protect\citeauthoryear{Rajeswaran, Kumar, Gupta, Vezzani, Schulman,
  Todorov, and Levine}{Rajeswaran et~al\mbox{.}}{2017}]%
        {rajeswaran2017learning}
\bibfield{author}{\bibinfo{person}{Aravind Rajeswaran}, \bibinfo{person}{Vikash
  Kumar}, \bibinfo{person}{Abhishek Gupta}, \bibinfo{person}{Giulia Vezzani},
  \bibinfo{person}{John Schulman}, \bibinfo{person}{Emanuel Todorov}, {and}
  \bibinfo{person}{Sergey Levine}.} \bibinfo{year}{2017}\natexlab{}.
\newblock \showarticletitle{Learning complex dexterous manipulation with deep
  reinforcement learning and demonstrations}.
\newblock \bibinfo{journal}{\emph{arXiv preprint arXiv:1709.10087}}
  (\bibinfo{year}{2017}).
\newblock


\bibitem[\protect\citeauthoryear{Roa and Su{\'a}rez}{Roa and
  Su{\'a}rez}{2009}]%
        {roa2009computation}
\bibfield{author}{\bibinfo{person}{M{\'a}ximo~A Roa} {and}
  \bibinfo{person}{Ra{\'u}l Su{\'a}rez}.} \bibinfo{year}{2009}\natexlab{}.
\newblock \showarticletitle{Computation of independent contact regions for
  grasping 3-d objects}.
\newblock \bibinfo{journal}{\emph{IEEE Transactions on Robotics}}
  \bibinfo{volume}{25}, \bibinfo{number}{4} (\bibinfo{year}{2009}),
  \bibinfo{pages}{839--850}.
\newblock


\bibitem[\protect\citeauthoryear{Sahbani, El-Khoury, and Bidaud}{Sahbani
  et~al\mbox{.}}{2012}]%
        {sahbani2012overview}
\bibfield{author}{\bibinfo{person}{Anis Sahbani}, \bibinfo{person}{Sahar
  El-Khoury}, {and} \bibinfo{person}{Philippe Bidaud}.}
  \bibinfo{year}{2012}\natexlab{}.
\newblock \showarticletitle{An overview of 3D object grasp synthesis
  algorithms}.
\newblock \bibinfo{journal}{\emph{Robotics and Autonomous Systems}}
  \bibinfo{volume}{60}, \bibinfo{number}{3} (\bibinfo{year}{2012}),
  \bibinfo{pages}{326--336}.
\newblock


\bibitem[\protect\citeauthoryear{Saxena, Driemeyer, Kearns, and Ng}{Saxena
  et~al\mbox{.}}{2007}]%
        {saxena2007robotic}
\bibfield{author}{\bibinfo{person}{Ashutosh Saxena}, \bibinfo{person}{Justin
  Driemeyer}, \bibinfo{person}{Justin Kearns}, {and} \bibinfo{person}{Andrew~Y
  Ng}.} \bibinfo{year}{2007}\natexlab{}.
\newblock \showarticletitle{Robotic grasping of novel objects}. In
  \bibinfo{booktitle}{\emph{Advances in neural information processing
  systems}}. \bibinfo{pages}{1209--1216}.
\newblock


\bibitem[\protect\citeauthoryear{Saxena, Wong, Quigley, and Ng}{Saxena
  et~al\mbox{.}}{2010}]%
        {saxena2010vision}
\bibfield{author}{\bibinfo{person}{Ashutosh Saxena}, \bibinfo{person}{Lawson
  Wong}, \bibinfo{person}{Morgan Quigley}, {and} \bibinfo{person}{Andrew~Y
  Ng}.} \bibinfo{year}{2010}\natexlab{}.
\newblock \showarticletitle{A vision-based system for grasping novel objects in
  cluttered environments}.
\newblock In \bibinfo{booktitle}{\emph{Robotics research}}.
  \bibinfo{publisher}{Springer}, \bibinfo{pages}{337--348}.
\newblock


\bibitem[\protect\citeauthoryear{Singh, Sha, Narayan, Achim, and Abbeel}{Singh
  et~al\mbox{.}}{2014}]%
        {singh2014bigbird}
\bibfield{author}{\bibinfo{person}{Arjun Singh}, \bibinfo{person}{James Sha},
  \bibinfo{person}{Karthik~S Narayan}, \bibinfo{person}{Tudor Achim}, {and}
  \bibinfo{person}{Pieter Abbeel}.} \bibinfo{year}{2014}\natexlab{}.
\newblock \showarticletitle{Bigbird: A large-scale 3d database of object
  instances}. In \bibinfo{booktitle}{\emph{2014 IEEE international conference
  on robotics and automation (ICRA)}}. IEEE, \bibinfo{pages}{509--516}.
\newblock


\bibitem[\protect\citeauthoryear{Song, Zeng, Lee, and Funkhouser}{Song
  et~al\mbox{.}}{2020}]%
        {song2020grasping}
\bibfield{author}{\bibinfo{person}{Shuran Song}, \bibinfo{person}{Andy Zeng},
  \bibinfo{person}{Johnny Lee}, {and} \bibinfo{person}{Thomas Funkhouser}.}
  \bibinfo{year}{2020}\natexlab{}.
\newblock \showarticletitle{Grasping in the wild: Learning 6dof closed-loop
  grasping from low-cost demonstrations}.
\newblock \bibinfo{journal}{\emph{IEEE Robotics and Automation Letters}}
  \bibinfo{volume}{5}, \bibinfo{number}{3} (\bibinfo{year}{2020}),
  \bibinfo{pages}{4978--4985}.
\newblock


\bibitem[\protect\citeauthoryear{Starke, Zhang, Komura, and Saito}{Starke
  et~al\mbox{.}}{2019}]%
        {starke2019neural}
\bibfield{author}{\bibinfo{person}{Sebastian Starke}, \bibinfo{person}{He
  Zhang}, \bibinfo{person}{Taku Komura}, {and} \bibinfo{person}{Jun Saito}.}
  \bibinfo{year}{2019}\natexlab{}.
\newblock \showarticletitle{Neural state machine for character-scene
  interactions.}
\newblock \bibinfo{journal}{\emph{ACM Trans. Graph.}} \bibinfo{volume}{38},
  \bibinfo{number}{6} (\bibinfo{year}{2019}), \bibinfo{pages}{209--1}.
\newblock


\bibitem[\protect\citeauthoryear{Sucan, Moll, and Kavraki}{Sucan
  et~al\mbox{.}}{2012}]%
        {sucan2012open}
\bibfield{author}{\bibinfo{person}{Ioan~A Sucan}, \bibinfo{person}{Mark Moll},
  {and} \bibinfo{person}{Lydia~E Kavraki}.} \bibinfo{year}{2012}\natexlab{}.
\newblock \showarticletitle{The open motion planning library}.
\newblock \bibinfo{journal}{\emph{IEEE Robotics \& Automation Magazine}}
  \bibinfo{volume}{19}, \bibinfo{number}{4} (\bibinfo{year}{2012}),
  \bibinfo{pages}{72--82}.
\newblock


\bibitem[\protect\citeauthoryear{Sutton and Barto}{Sutton and Barto}{2018}]%
        {sutton2018reinforcement}
\bibfield{author}{\bibinfo{person}{Richard~S Sutton} {and}
  \bibinfo{person}{Andrew~G Barto}.} \bibinfo{year}{2018}\natexlab{}.
\newblock \bibinfo{booktitle}{\emph{Reinforcement learning: An introduction}}.
\newblock \bibinfo{publisher}{MIT press}.
\newblock


\bibitem[\protect\citeauthoryear{Van~der Merwe, Lu, Sundaralingam, Matak, and
  Hermans}{Van~der Merwe et~al\mbox{.}}{2019}]%
        {van2019learning}
\bibfield{author}{\bibinfo{person}{Mark Van~der Merwe},
  \bibinfo{person}{Qingkai Lu}, \bibinfo{person}{Balakumar Sundaralingam},
  \bibinfo{person}{Martin Matak}, {and} \bibinfo{person}{Tucker Hermans}.}
  \bibinfo{year}{2019}\natexlab{}.
\newblock \showarticletitle{Learning Continuous 3D Reconstructions for
  Geometrically Aware Grasping}.
\newblock \bibinfo{journal}{\emph{arXiv preprint arXiv:1910.00983}}
  (\bibinfo{year}{2019}).
\newblock


\bibitem[\protect\citeauthoryear{Varley, DeChant, Richardson, Ruales, and
  Allen}{Varley et~al\mbox{.}}{2017}]%
        {varley2017shape}
\bibfield{author}{\bibinfo{person}{Jacob Varley}, \bibinfo{person}{Chad
  DeChant}, \bibinfo{person}{Adam Richardson}, \bibinfo{person}{Joaqu{\'\i}n
  Ruales}, {and} \bibinfo{person}{Peter Allen}.}
  \bibinfo{year}{2017}\natexlab{}.
\newblock \showarticletitle{Shape completion enabled robotic grasping}. In
  \bibinfo{booktitle}{\emph{2017 IEEE/RSJ international conference on
  intelligent robots and systems (IROS)}}. IEEE, \bibinfo{pages}{2442--2447}.
\newblock


\bibitem[\protect\citeauthoryear{Vecerik, Hester, Scholz, Wang, Pietquin, Piot,
  Heess, Roth{\"o}rl, Lampe, and Riedmiller}{Vecerik et~al\mbox{.}}{2017}]%
        {vecerik2017leveraging}
\bibfield{author}{\bibinfo{person}{Mel Vecerik}, \bibinfo{person}{Todd Hester},
  \bibinfo{person}{Jonathan Scholz}, \bibinfo{person}{Fumin Wang},
  \bibinfo{person}{Olivier Pietquin}, \bibinfo{person}{Bilal Piot},
  \bibinfo{person}{Nicolas Heess}, \bibinfo{person}{Thomas Roth{\"o}rl},
  \bibinfo{person}{Thomas Lampe}, {and} \bibinfo{person}{Martin Riedmiller}.}
  \bibinfo{year}{2017}\natexlab{}.
\newblock \showarticletitle{Leveraging demonstrations for deep reinforcement
  learning on robotics problems with sparse rewards}.
\newblock \bibinfo{journal}{\emph{arXiv preprint arXiv:1707.08817}}
  (\bibinfo{year}{2017}).
\newblock


\bibitem[\protect\citeauthoryear{Viereck, Pas, Saenko, and Platt}{Viereck
  et~al\mbox{.}}{2017}]%
        {viereck2017learning}
\bibfield{author}{\bibinfo{person}{Ulrich Viereck}, \bibinfo{person}{Andreas
  Pas}, \bibinfo{person}{Kate Saenko}, {and} \bibinfo{person}{Robert Platt}.}
  \bibinfo{year}{2017}\natexlab{}.
\newblock \showarticletitle{Learning a visuomotor controller for real world
  robotic grasping using simulated depth images}. In
  \bibinfo{booktitle}{\emph{Conference on Robot Learning}}. PMLR,
  \bibinfo{pages}{291--300}.
\newblock


\bibitem[\protect\citeauthoryear{Wang, Xiang, and Fox}{Wang
  et~al\mbox{.}}{2019}]%
        {wang2019manipulation}
\bibfield{author}{\bibinfo{person}{Lirui Wang}, \bibinfo{person}{Yu Xiang},
  {and} \bibinfo{person}{Dieter Fox}.} \bibinfo{year}{2019}\natexlab{}.
\newblock \showarticletitle{Manipulation trajectory optimization with online
  grasp synthesis and selection}.
\newblock \bibinfo{journal}{\emph{arXiv preprint arXiv:1911.10280}}
  (\bibinfo{year}{2019}).
\newblock


\bibitem[\protect\citeauthoryear{Wohlkinger, Aldoma, Rusu, and
  Vincze}{Wohlkinger et~al\mbox{.}}{2012}]%
        {wohlkinger20123dnet}
\bibfield{author}{\bibinfo{person}{Walter Wohlkinger}, \bibinfo{person}{Aitor
  Aldoma}, \bibinfo{person}{Radu~B Rusu}, {and} \bibinfo{person}{Markus
  Vincze}.} \bibinfo{year}{2012}\natexlab{}.
\newblock \showarticletitle{3dnet: Large-scale object class recognition from
  cad models}. In \bibinfo{booktitle}{\emph{2012 IEEE international conference
  on robotics and automation}}. IEEE, \bibinfo{pages}{5384--5391}.
\newblock


\bibitem[\protect\citeauthoryear{Xu, Qi, Agrawal, and Song}{Xu
  et~al\mbox{.}}{2021}]%
        {xu2021adagrasp}
\bibfield{author}{\bibinfo{person}{Zhenjia Xu}, \bibinfo{person}{Beichun Qi},
  \bibinfo{person}{Shubham Agrawal}, {and} \bibinfo{person}{Shuran Song}.}
  \bibinfo{year}{2021}\natexlab{}.
\newblock \showarticletitle{Adagrasp: Learning an adaptive gripper-aware
  grasping policy}. In \bibinfo{booktitle}{\emph{2021 IEEE International
  Conference on Robotics and Automation (ICRA)}}. IEEE,
  \bibinfo{pages}{4620--4626}.
\newblock


\bibitem[\protect\citeauthoryear{Zhao, Choi, and Komura}{Zhao
  et~al\mbox{.}}{2017}]%
        {zhao2017character}
\bibfield{author}{\bibinfo{person}{Xi Zhao}, \bibinfo{person}{Myung~Geol Choi},
  {and} \bibinfo{person}{Taku Komura}.} \bibinfo{year}{2017}\natexlab{}.
\newblock \showarticletitle{Character-object interaction retrieval using the
  interaction bisector surface}.
\newblock \bibinfo{journal}{\emph{Computer Graphics Forum (Proc.
  Eurographics)}} \bibinfo{volume}{36}, \bibinfo{number}{2}
  (\bibinfo{year}{2017}), \bibinfo{pages}{119--129}.
\newblock


\bibitem[\protect\citeauthoryear{Zhao, Wang, and Komura}{Zhao
  et~al\mbox{.}}{2014}]%
        {zhao2014IBS}
\bibfield{author}{\bibinfo{person}{Xi Zhao}, \bibinfo{person}{He Wang}, {and}
  \bibinfo{person}{Taku Komura}.} \bibinfo{year}{2014}\natexlab{}.
\newblock \showarticletitle{Indexing 3d scenes using the interaction bisector
  surface}.
\newblock \bibinfo{journal}{\emph{ACM Trans. on Graphics}}
  \bibinfo{volume}{33}, \bibinfo{number}{3} (\bibinfo{year}{2014}),
  \bibinfo{pages}{1--14}.
\newblock


\end{thebibliography}

\end{document}